\documentclass[10pt]{article} 
\usepackage[dvipsnames]{xcolor}
\usepackage[preprint]{rlc}

\usepackage{mathtools}          
\usepackage{mathrsfs}           
\mathtoolsset{showonlyrefs}     
\usepackage[space]{grffile}     

\usepackage[utf8]{inputenc} 
\usepackage[T1]{fontenc}    
\usepackage{hyperref}       
\usepackage{url}            
\usepackage{booktabs}       
\usepackage{amssymb}
\usepackage{amsfonts}       
\usepackage{amsmath}        
\usepackage{algorithm}
\usepackage{algpseudocode}
\usepackage{nicefrac}       
\usepackage{microtype}      
\usepackage{lipsum}
\usepackage{fancyhdr}       
\usepackage{graphicx}       
\usepackage{xpatch}
\graphicspath{{media/}}     
\usepackage{svg}
\usepackage{orcidlink}
\usepackage{soul}
\usepackage{multirow}
\usepackage{array}
\usepackage{hwemoji}
\usepackage{subcaption}
\usepackage{natbib}
\usepackage{xspace}

\newcommand{\blindurl}[1]{\url{#1}}
\newcommand{\blindhref}[2]{\href{#1}{#2}}
\newcommand{\blindhreftiny}[2]{\href{#1}{#2}}

\newcommand{\srl}{\textsc{SnapshotRL}\xspace}

\colorlet{LightAquamarine}{Aquamarine!40!White}
\colorlet{LightSalmon}{Salmon!40!White}
\colorlet{LightGoldenrod}{Goldenrod!40!White}
\colorlet{LightOrchid}{Orchid!40!White}
\colorlet{LightYellow}{Yellow!30!White}

\title{Snapshot Reinforcement Learning: Leveraging Prior Trajectories for Efficiency}

\author{\name{Yanxiao Zhao}$^{1, 2}$ \orcidlink{0000-0001-9842-4706}\\
  \email{zhaoyanxiao21@mails.ucas.ac.cn}
  \And
  \name{Yangge Qian}$^{1, 2}$ \orcidlink{0009-0003-1787-7108}\\
  \email{qianyangge20@mails.ucas.ac.cn}\\
  \And
  \name{Tianyi Wang}$^{1, 2}$\\
  \email{wangtianyi22@mails.ucas.ac.cn}\\
  \And
  \name{Jingyang Shan}$^{1, 2}$\\
  \email{shanjingyang21@mails.ucas.ac.cn}\\
  \And
  \name{Xiaolin Qin}$^{1, 2}$\thanks{Corresponding author}\\
  \email{qinxl2001@126.com}\\
  \AND
  \addr{$^{1}$ Chengdu Institute of Computer Applications, Chinese Academy of Sciences}\\
  \addr{$^{2}$ School of Computer Science and Technology, University of Chinese Academy of Sciences}
}

\begin{document}


\maketitle

\begin{abstract}
    Deep reinforcement learning (DRL) algorithms require substantial samples and computational resources to achieve higher performance, which restricts their practical application and poses challenges for further development. 
    Given the constraint of limited resources, it is essential to leverage existing computational work (e.g., learned policies, samples) to enhance sample efficiency and reduce the computational resource consumption of DRL algorithms. 
    Previous works to leverage existing computational work require intrusive modifications to existing algorithms and models, designed specifically for specific algorithms, lacking flexibility and universality.
    In this paper, we present the Snapshot Reinforcement Learning (\srl) framework, which enhances sample efficiency by simply altering environments, without making any modifications to algorithms and models.
    By allowing student agents to choose states in teacher trajectories as the initial state to sample, \srl can effectively utilize teacher trajectories to assist student agents in training, allowing student agents to explore a larger state space at the early training phase.
    We propose a simple and effective \srl baseline algorithm, S3RL, which integrates well with existing DRL algorithms.
    Our experiments demonstrate that integrating S3RL with TD3, SAC, and PPO algorithms on the MuJoCo benchmark significantly improves sample efficiency and average return, without extra samples and additional computational resources.
\end{abstract}

\section{Introduction}\label{sec:introduction}

Deep Reinforcement Learning (DRL) has enjoyed numerous accomplishments in game, simulation, and real-world environments.
However, the development of powerful agents requires a significant amount of samples and computational resources.
For example, AlphaStar~\citep{vinyals2019grandmaster} was trained using 16 TPU-v3 for 14 days, during which each agent used the equivalent of 200 years of the real-time StarCraft II game.
Similarly, Robotic Transformer 2 (RT-2)~\citep{rt22023arxiv} utilized demonstration data collected by 13 robots over 17 months in an office kitchen environment.
This obstacle prevents researchers who lack necessary resources from reproducing these works, thus limiting the applications and development of these works.

\begin{figure}[ht]
    \centering
    \includegraphics[width=1.0\linewidth]{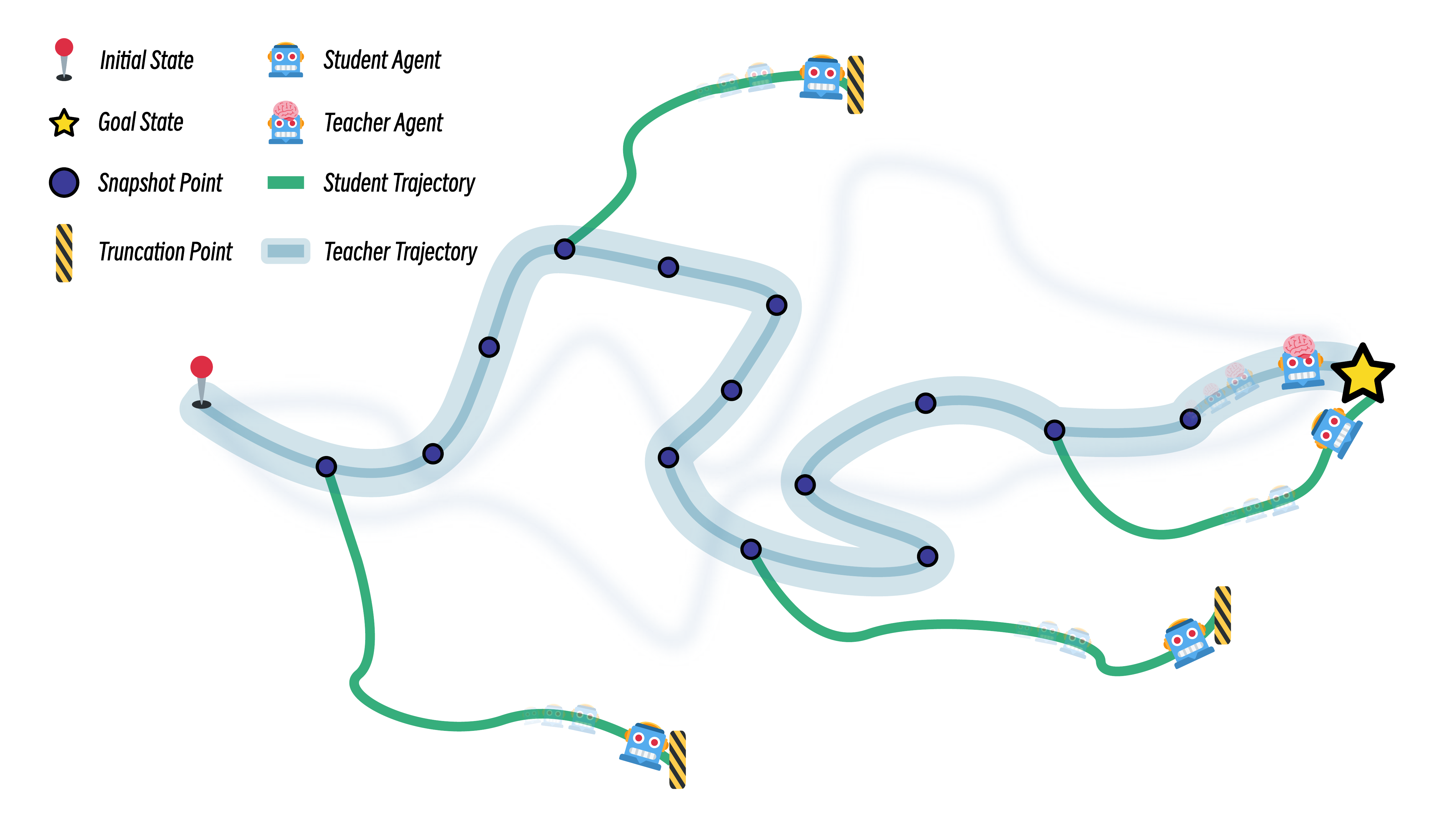}
    \caption{
Schematic of S3RL training process. 
The figure illustrates a teacher trajectory (light blue line with outline) from the initial point (red pin) to the goal point (yellow pentagram).
Dark blue dots scattered on this trajectory indicate environment snapshots obtained from the teacher agent's interaction with environment, from which the student agent starts new training represented by green trajectories.
Truncation points(black and yellow squares) on the right of three student trajectories signify truncated training implemented to prevent the student agent from deviating excessively from the teacher trajectory.
The student trajectory on far right reaches the goal point, demonstrating that the student agent can successfully accomplish tasks.
The figure vividly portrays the mechanism and objective of S3RL: to support the training of new agents effectively by leveraging environment snapshots.
    }\label{fig:srl}
\end{figure}

In light of this, Reincarnating Reinforcement Learning (RRL)~\citep{DBLP:conf/nips/AgarwalSCCB22} emerges as a promising research workflow.
RRL aims to maximize the utilization of pre-existing computational work, thus releasing researchers from the need for tabula rasa when training agents and ultimately enhancing sample efficiency and reducing computational resource consumption. 
Previous RRL studies mainly concentrated on reusing pre-existing agent models or replay buffers, to enhance the performance of new agents.
For example, seminal works such as those by \citet{DBLP:journals/corr/VecerikHSWPPHRL17, DBLP:journals/corr/abs-2006-09359, DBLP:journals/corr/abs-2111-05424, wu2022supported, nakamoto2023calql, DBLP:journals/corr/abs-2303-17396} have capitalized on leveraging previously gathered demonstration data for offline pre-training, followed by careful online fine-tuning to refine agent behaviors.
In parallel, \citet{DBLP:conf/icml/PardoTLK18, DBLP:journals/jmlr/RossGB11, DBLP:conf/nips/AgarwalSCCB22} combined with prior agents to propose special loss functions.
Further, \citet{DBLP:conf/aistats/CzarneckiPOJSJ19, DBLP:conf/iclr/0002BB18, DBLP:journals/corr/abs-2301-12876} utilize Q-function of teacher agent to compute additional rewards, guiding learning process of student agent.

However, these works usually require intrusive modifications to existing algorithms and models.
Such modifications are designed for specific algorithms, lacking flexibility and universality.
Researchers need to frequently adjust the design of algorithms and models during experimental studies to verify their ideas.
Integrating their newly designed RL algorithms with existing RRL strategies again creates additional workloads, which hardly meet their needs.

We have dubbed our framework Snapshot Reinforcement Learning (\srl).
\srl can enhance sample efficiency by simply altering environments, without making any modifications to algorithms and models.
For simulated environments, the implementation of \srl merely involves incorporating wrappers that enable the loading of snapshots into the environment, thus avoiding the necessity for extensive code modifications and significantly easing its integration into various RL research works.
Environment snapshots preserve complete data of the simulation environment and allow the environment to be restored to a specific previously saved snapshot point.
Our main idea is that using snapshots from teacher agent trajectories to assist student agent training allows student agents to choose states within teacher agent trajectories as initial points to begin sampling, leading to a broader exploration of states by student agents during the early training phase.
By training with snapshots generated by teacher agents with environment, our framework can effectively leverage the experience accumulated by teacher agents, similar to the practice of endgame training in the game of Go.

In this paper, we first introduce \srl research framework, propose standardized evaluation suggestions, and analyze the challenges faced by this framework.
Subsequently, we designed and introduced \srl with Status Classification and Student Trajectory Truncation (S3RL), a simple and effective \srl baseline algorithm developed for these challenges.
The schematic process of S3RL training is illustrated in Figure~\ref{fig:srl}.
Our experimental results show that, on the Gymnasium MuJoCo benchmark~\citep{todorov2012mujoco, towers_gymnasium_2023}, when integrated with TD3, SAC, and PPO algorithms, S3RL achieves superior performance over baseline methods with just 50\% of timesteps, significant improvements sample efficiency.
It is important to note that the performance improvement with S3RL was achieved without increasing any computational cost in the learning part and without directly providing additional samples to student agents, which is different from previous RRL works.

\section{Preliminaries}\label{sec:preliminaries}

In our RL framework, we consider a discrete-time, stochastic, and Markov Decision Process (MDP). This process is defined by a tuple $(S, A, P, R, O, \gamma)$, where

\begin{itemize}
    \item $S$ is the state space, which represents different states of the system.
    \item $A$ is the action space, which includes all possible actions that can be taken by the agent.
    \item $P : S \times A \times S \rightarrow [0,1]$ is the state-transition probability function. It quantifies the likelihood of transitioning from one state to another, given a particular action.
    \item $R : S \times A \times S \rightarrow \mathbb{R}$ is the reward function. $R(s,a,s^\prime)$ denotes the immediate reward received after transitioning from state $s$ to state $s^\prime$, due to action $a$.
    \item $O$ is the observation space, represented by a function $O : S \rightarrow \mathbb{O}$, where $\mathbb{O}$ is the set of all possible observations. $O(s)$ denotes the observation when the system is in state $s$.
    \item $\gamma$ is a discount factor. $\gamma \in [0, 1)$
\end{itemize}

Additionally, we introduce the concept of environment snapshot, which is an extended representation of environment at a certain timestep.
An environment snapshot captures not only the current state of the system but also the complete set of parameters defining the MDP.
This allows for the possibility of preserving the entire state of the system, including the MDP configuration, facilitating operations such as environment resets to a past state.
We denote an environment snapshot as follows:

$$
\mathcal{S}_i = \langle s_i \mid (S, A, P, R, O, \gamma) \rangle
$$

Here, $\mathcal{S}_i$ includes the current state $s_i$ and the tuple representing the entire MDP configuration.

An agent's purpose in this model is to learn a policy $\pi: S \rightarrow A$, which selects an action $a = \pi(s)$ to execute in state $s$. The aim is to maximize the expected cumulative reward over time.

\section{\srl: A Framework for Leveraging Prior Trajectories}\label{sec:srl-a-framework-for-leveraging-prior-trajectories}

In this section, we introduce a new framework for enhanced sample efficiency in RL algorithms — \srl.
We elucidate the core mechanism of \srl, including how to capture and store trajectory snapshots, the principles for selecting and applying snapshots, as well as the intuition and expected outcomes behind this mechanism.

For the most straightforward implementation of the \srl framework in pseudocode, please refer to the parts of Algorithm~\ref{alg:s3rl} excluding those marked by \includegraphics[height=1.0em]{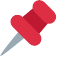}.

\subsection{Procuring Snapshots}\label{ssec:procuring-snapshots}

We found that \srl is highly sensitive to the distribution of snapshots, and this distribution directly impacts algorithm performance.
Conventionally, algorithms that leverage environment snapshots tend to manually select snapshots deemed important by experts, a practice that is both random and inflexible, making it difficult to compare and evaluate the performance of different \srl algorithms.
To standardize research on \srl, we systematically save agent models, and during each student agent training process, we interactively generate multiple trajectories with the environment, saving the environment snapshot of each step in the snapshot collection $\mathcal{D}_{\mathcal{S}}$ for further use.

Our design facilitates the flexible creation of new \srl algorithms by researchers, who have access to a wealth of information, including Q-values output by agent models. Varying the random seed generates different collections of $\mathcal{D}_{\mathcal{S}}$, which helps us to evaluate our new algorithms more accurately.

\subsection{Weaning off Snapshots}\label{ssec:weaning-off-snapshots}

The goal of \srl is to enhance the sample efficiency of existing reinforcement learning algorithms on existing environments, rather than create environments inherently more favorable for agent training.
During training process of \srl algorithms, the environment used for training is different from the one used for evaluation, and the state distribution of training environment is actively controlled.
The ultimate goal is for agents to adapt and perform better on the original environment, a transition that involves progressive reduction of dependence on snapshots.
In the algorithm we present later, \srl is applied only during the first $10\%$ of training timesteps, after which the agent continues training in the unaltered, original environment.
Our results indicate that using \srl only in the initial training phase significantly improves sample efficiency of existing algorithms.


\section{S3RL: A simple \srl baseline}\label{sec:s3rl-a-simple-srl-baseline}

Following the introduction and analysis of \srl in the previous section, this section will present \srl with Status Classification and Student Trajectory Truncation (S3RL), a baseline algorithm for \srl.
S3RL consists of two improvement parts: (1) Status Classification (SC) and (2) Student Trajectory Truncation (STT), which are designed to address the challenges of state duplication and insufficient influence within \srl.
Please refer to Algorithm~\ref{alg:s3rl} for the pseudocode of S3RL.

\begin{algorithm}
  \small
  \caption{S3RL: \colorbox{LightSalmon}{SnapshotRL} with \colorbox{LightAquamarine}{Status Classification} and \colorbox{LightOrchid}{Student Trajectory Truncation}}\label{alg:s3rl}
  \begin{algorithmic}[1]
  \State\textbf{Input:} a environment $E$, a collection of environment snapshots $\{(\mathcal{S}_1, q_1), (\mathcal{S}_2, q_2), \cdots, (\mathcal{S}_N, q_N)\}$, maximum length of student agent trajectory $T$, a RL algorithm $Alg$ such as TD3.
  \State Initialize policy $\pi$ from scratch. Initialize snapshot dataset $\mathcal{D}_{\mathcal{S}}\gets \{(\mathcal{S}_1, q_1), (\mathcal{S}_2, q_2), \cdots, (\mathcal{S}_N, q_N)\}$.
  \State \colorbox{LightAquamarine}{$\mathcal{D}_{\mathcal{S}}^\prime \gets \Call{Kmeans}{\mathcal{D}_{\mathcal{S}}}$\includegraphics[height=1.0em]{media/emoji_pin.pdf}}
  \Statex After applying K-means, partition $\mathcal{D}_{\mathcal{S}}$ into $k$ disjoint clusters by Q-value, with each cluster $C_i$ containing $n_i$ states.
  \Statex $\mathcal{D}_{\mathcal{S}}^\prime = \{C_1, C_2, \ldots, C_k\} \text{ where } C_i = \{\mathcal{S}_{i,1}, \mathcal{S}_{i,2}, \ldots, \mathcal{S}_{i,n_i}\}, \sum_{1}^{k}n_i = N.$
  \While{$number\_of\_iterations \leq max\_iterations$}
      \State $E$ = \Call{Reset}{$E$}
      \If{in snapshot environment train phase}
          \State \colorbox{LightSalmon}{$C$ = \Call{RandomChoice}{$\mathcal{D}_{\mathcal{S}}^\prime$}}
          \State \colorbox{LightSalmon}{$\mathcal{S}$ = \Call{RandomChoice}{$C$}}
          \State \colorbox{LightSalmon}{$E$ = \Call{LoadSnapshot}{$E,\mathcal{S}$}}
      \EndIf
      \State Roll out policy $\pi$ within the environment $E$ using the exploration method $Alg$ to get \colorbox{LightOrchid}{a time-limited} \colorbox{LightOrchid}{trajectory $\{(o_1, a_1, r_1), \cdots, (o_{t}, a_{t}, r_{t})\}$\includegraphics[height=1.0em]{media/emoji_pin.pdf}}, where the length of the trajectory, indicated by $t$, will not exceed the predefined limit $T$.
      \State Update $\pi$ by $Alg$.
  \EndWhile
  \end{algorithmic}
\end{algorithm}

\subsection{Status Classification}\label{ssec:status-classification}

Within our \srl algorithm, we identified an issue: the snapshot collection $\mathcal{D}_{\mathcal{S}}$ often contains many duplicate or similar snapshots, resulting in an excessively high likelihood of selecting similar snapshots during random sampling processes.
Taking the MuJoCo Hopper environment\footnote{Documentation for Hopper Environment: \href{https://gymnasium.farama.org/environments/mujoco/hopper/}{https://gymnasium.farama.org/environments/mujoco/hopper/}} as an example, a well-trained monopedal robot quickly enters a phase of motion marked by distinctive periodic characteristics after it has started.
If randomly selected from all snapshots without adjustment, it might focus too much on the periodic phase, neglecting crucial snapshots like those found during the start-up phase.

To address this issue, we have developed a state classification strategy, which is based on Q-value of state in snapshot.
Using the standard K-means clustering algorithm, we categorize snapshot according to the Q-values produced by teacher agent and uniformly select snapshot from each category to ensure balanced category coverage.
Our work does not delve into which specific snapshots are most conducive to the learning process of student agents.
We simply propose a straightforward method of state classification designed to maintain an equilibrium in the significance attributed to various snapshots.

\subsection{Student Trajectory Truncation}\label{ssec:student-trajectory-truncation}

In \srl framework, only initial states of student agent trajectories is regulated.
However, such an approach might not be sufficient for tasks that require long-term foresight.
The influence of initial states tends to decrease as student agent trajectories lengthen.
This is particularly evident in the early stages of training, when student agents may inadvertently fall into adverse states, quickly diminishing the effect of \srl.

To address this challenge, we propose Student Trajectory Truncation (STT) strategy. STT prematurely truncates student agent trajectories (e.g., setting the maximum episode length in MuJoCo environments to $100$ instead of the default $1000$ steps).
This strategy increases the frequency with which student agents encounter states within $\mathcal{D}_{\mathcal{S}}$, aiming to enhance the agent's learning opportunities from the initial states that are controlled by \srl.

\section{Experiments}\label{sec:experiments}

Our experiments will answer the following questions:
(1) How does \srl affect the learned policies quality?
(2) What are the most key components of \srl?
(3) Does \srl have strong robustness and algorithmic compatibility?

We first train five teacher agents using CleanRL's TD3 implementation, each for 1 million timesteps on MuJoCo benchmark, with five random seeds.
Teacher models can be found in Table~\ref{tab:td3-models}.
We select the best performing teacher agent for generating snapshot dataset.
To ensure the robustness of our experimental results, we generate a unique snapshot dataset for each run using a teacher agent with varying random seeds.
The teacher agent interacts with the environment for ten episodes and saves an environment snapshot into a snapshot dataset every ten timesteps.

\begin{figure}
    \centering
    \includegraphics[width=1.0\linewidth]{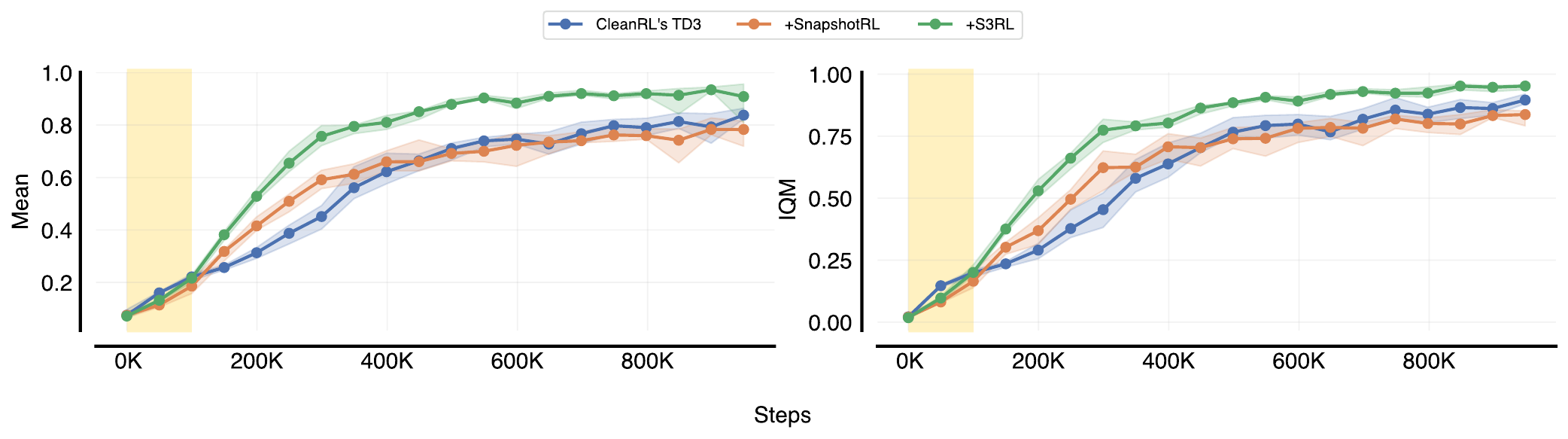}
    \caption{
        Learning curves sample efficiency comparison of TD3, \srl{}+TD3, and S3RL+TD3 on six MuJoCo environments. For individual environment results, see Figure~\ref{fig:td3_indiv}.
    }\label{fig:td3}
\end{figure}

Subsequently, we integrate \srl and S3RL with TD3 and run it on six MuJoCo environments, including \texttt{Hopper-v4}, \texttt{Walker2d-v4}, \texttt{HalfCheetah-v4}, \texttt{Ant-v4}, \texttt{Swimmer-v4}, and \texttt{Humanoid-v4}.
Our experimental results are shown in Figure~\ref{fig:td3} and~\ref{fig:td3_indiv}.
We use \srl training only for the first $100,000$ timesteps, after that we use the original environment for training, the highlighted part in figures is \colorbox{LightYellow}{\srl training phase}.
The results show that the TD3 algorithm using only \srl cannot achieve better performance than TD3, and even performs worse in some environments.
However, when we combine SC and STT strategies with \srl, sample efficiency and average return of TD3 are significantly improved in all six environments.
We also evaluated the performance of S3RL+TD3 under different levels of teacher agents, see Appendix~\ref{ssec:sweep-of-teacher-models}.

To evaluate the compatibility of the S3RL algorithm, we conducted a series of experiments integrating S3RL with SAC and PPO algorithms.
For detailed information about these experiments, please refer to Appendices~\ref{ssec:evaluating-s3rl-with-soft-actor-critic} and~\ref{ssec:evaluating-s3rl-with-proximal-policy-optimization}.
Our results indicate that while S3RL significantly enhances the performance when combined with off-policy algorithms like TD3 and SAC, the performance improvements with the on-policy PPO algorithm are comparatively modest.
See Appendix~\ref{ssec:evaluating-s3rl-with-proximal-policy-optimization} for analysis and discussion of this phenomenon.

\subsection{Ablation Study}\label{ssec:ablation-study}

\begin{figure}[ht]
    \centering
    \includegraphics[width=1.0\linewidth]{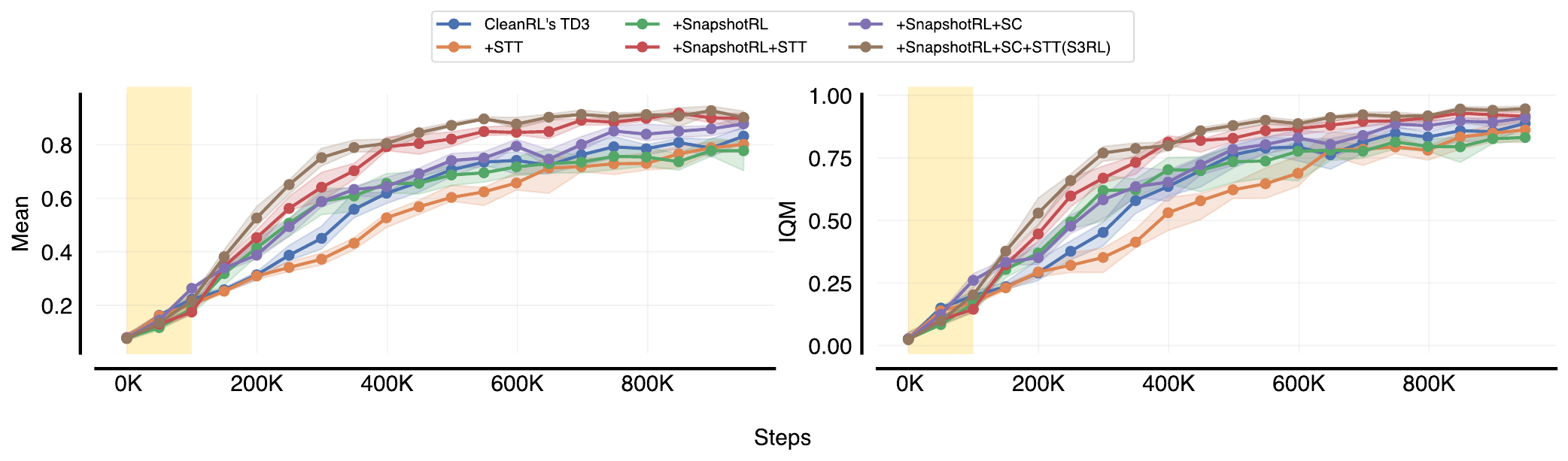}
    \caption{
        Ablation study results showing the impact of key components on the sample efficiency of S3RL+TD3 on six MuJoCo environments. For individual environment results, see Figure~\ref{fig:td3_ablation_indiv}.
    }\label{fig:td3_ablation}
\end{figure}

We also conducted ablation experiments, and the results are shown in Figure~\ref{fig:td3_ablation}.
\srl{}+SC+STT (S3RL) significantly outperformed its ablation variants (\srl, \srl{}+SC and \srl{}+STT) in terms of both sample efficiency and average return.
This indicates that SC and STT methods are effective ways to improve the performance of \srl, and can improve the performance of \srl whether used alone or in combination.

In addition,~\citet{DBLP:conf/icml/PardoTLK18} pointed out that premature truncation can affect algorithm performance.
Our ablation experiment TD3+STT sets the truncation step to 100 steps in the first $100,000$ timesteps, which reverts to the default setting of $1000$ steps.
The results show that without \srl, STT has a negative impact on the performance of TD3.
This result indicates that the performance improvement does not come from the premature truncation effect of STT, but from the fact that STT enhances the impact of \srl on training.

\subsection{Hyperparameter Robustness Study}\label{ssec:hyperparameter-robustness-study}

\begin{figure}[ht]
    \centering
    \includegraphics[width=1.0\linewidth]{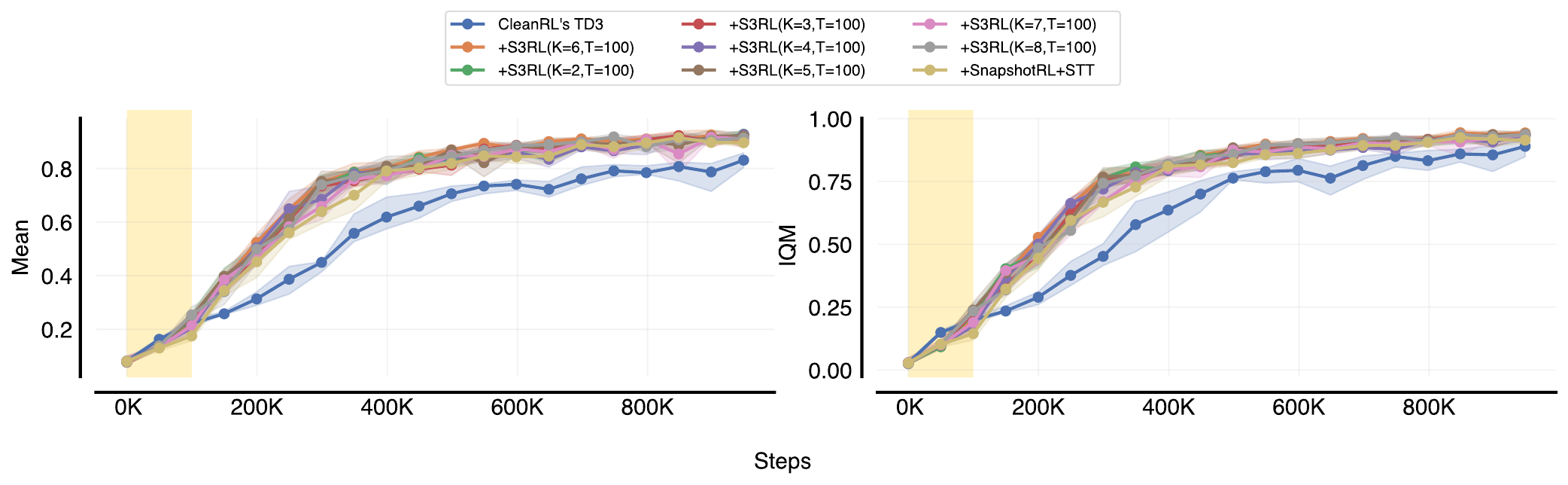}
    \caption{
        Learning curves sample efficiency sweeps for S3RL+TD3 across $K$ on six MuJoCo environments. For individual environment results, see Figure~\ref{fig:td3_sweep_k_indiv}.
    }\label{fig:td3_sweep_k}
\end{figure}

\begin{figure}[ht]
    \centering
    \includegraphics[width=1.0\linewidth]{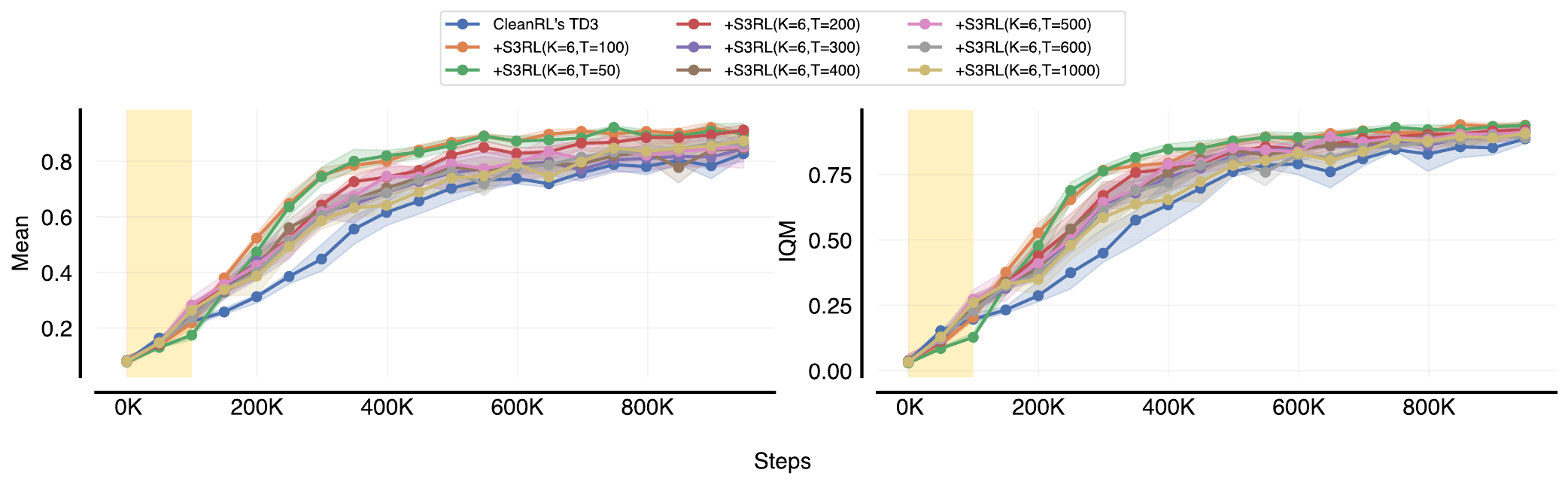}
    \caption{
        Learning curves sample efficiency sweeps for S3RL+TD3 across $T$ on six MuJoCo environments. For individual environment results, see Figure~\ref{fig:td3_sweep_t_indiv}.
    }\label{fig:td3_sweep_t}
\end{figure}

In S3RL, both SC and STT components possess a hyperparameter each, namely the number of clusters $K$ and the truncation step $T$, respectively.
To demonstrate that the performance improvements obtained with our algorithm are mainly attributable to its design innovations, rather than meticulous parameter optimization, we swept a range of hyperparameters, reporting algorithm performance under these varying conditions.
The experimental outcomes, as illustrated in Figure~\ref{fig:td3_sweep_k} and Figure~\ref{fig:td3_sweep_t}, reveal that our algorithm's performance is not critically dependent on the fine-tuning of the cluster count $K$, and that the truncation step $T$ exhibits a negative correlation with performance metrics within a certain range.

These findings bolster our confidence in S3RL: it can achieve performance gains through its innovative design while remaining robust to choices in hyperparameter settings.
Specifically, the results show that although the number of clusters $K$ minimally impacts performance, appropriate selection of the truncation step $T$ can further optimize performance outcomes.
This suggests that fine-tuning the truncation strategy could offer new avenues for improvements in the efficiency and effectiveness of the algorithm in the future.
In forthcoming work, we anticipate that adjusting these parameters through adaptive methods or employing advanced parameter search strategies could further enhance the performance of S3RL and streamline its application process.

\section{Related Work}\label{sec:related-work}

In this section, we provide an overview of representative related works in this field, offering a comparative analysis with our contributions.

\citet{DBLP:journals/corr/HosuR16, DBLP:journals/corr/abs-1812-03381, DBLP:journals/corr/abs-1810-10654, DBLP:conf/icra/NairMAZA18} use states in demonstration trajectories as initial states of agents, and demonstration trajectories are obtained by experts interacting with environment or planner solving it.
\citet{DBLP:journals/tog/PengALP18} uses a set of states carefully selected by human experts as a set of initial states of agent.
The cost of obtaining these demonstration data is relatively high. It depends on human experts, while our work only uses demonstration data obtained from previous interactions between agent and environment, which is easy to obtain and reproduce.
Our work focuses on using suboptimal demonstration data from prior agents, rather than expert demonstration data.
\citet{messikommer2023contrastive} saves the previously visited states during the training process, and uses states as initial states in subsequent training.
It proposes to use an Embedding Network to extract features from the previously visited states and then use these features to classify states.
Different from State Classification proposed in our work, we use Q-values output by prior agents as features required for classification.
We believe that prior agents have already learned some helpful information, which is reflected in Q-values, so we do not need to retrain an Embedding Network for classification.

A work similar to our work is Jump-Start Reinforcement Learning (JSRL)~\citep{DBLP:conf/icml/UchenduX0ZYSBFM23}, which first uses a teacher agent in each trajectory, and then uses a student agent to help the student agent explore.
JSRL puts the samples collected by teacher agent into the experience replay buffer.
SnapshotRL can be simply understood as JSRL without using teacher samples.
Although SnapshotRL and JSRL have similar working procedures, the fundamental difference between SnapshotRL and JSRL is that using prior agents is not necessary for SnapshotRL, which is just one way we standardize the research of SnapshotRL.
In specific cases, SnapshotRL can also use Snapshots obtained in other ways.

\section{Conclusion}\label{sec:conclusion}

The contributions of this paper are as follows.
(1) We have proposed \srl framework, which focuses on leveraging prior trajectories to enhance sample efficiency of new agents.
(2) We have designed S3RL, a baseline algorithm for \srl, which consists of two improvement parts, SC and STT, designed to address challenges of state duplication and insufficient influence within \srl.
(3) Experiments were carefully designed to analyze the utility of components of S3RL, assess its robustness, and the performance improvements of integrating S3RL with various RL algorithms.

In future work, we aim to further explore the potential of \srl, studying how \srl can be applied to more complex environments and real-world applications.
Additionally, we plan to study the integration of \srl with other methodologies, particularly those that leverage prior computational efforts, to ensure compatibility and more effective utilization.

\section{Limitation}\label{sec:limitation}

There are several limitations in this research.
Firstly, our method depends on trajectories provided by teacher agents.
Thus, its effectiveness might be limited in environments where teacher agents perform inadequately.
If teacher agents cannot provide high-quality demonstrations, this could impact the learning efficacy of student agents.
Secondly, our method requires environment snapshots, yet acquiring complete snapshots can be highly challenging, or restoring from a particular state might incur significant costs in some real-world environments.
Lastly, our method may have limitations when applied to on-policy algorithms.
Our experiments revealed that \srl{}+PPO and S3RL+PPO only exhibited satisfactory performance in a limited set of environments, and a detailed analysis of the reasons is provided in Appendix~\ref{ssec:evaluating-s3rl-with-proximal-policy-optimization}.

\section*{Reproducibility Statement}\label{sec:reproducibility-statement}

To enhance the reproducibility of our work and support the validation and further research by peers, we have provided a detailed description of our implementation in Section~\ref{sec:srl-a-framework-for-leveraging-prior-trajectories}, Section~\ref{sec:s3rl-a-simple-srl-baseline}, and Appendix~\ref{sec:experiment-details}, with hyperparameters and models listed in Appendices~\ref{sec:hyperparameter} and \ref{sec:teacher-models}, respectively.
All associated source code, models, and Weights \& Biases experiment reports are accessible via \blindhref{https://sdpkjc.github.io/snapshotrl}{sdpkjc.github.io/snapshotrl}.

Our experiment results are adapted for comparison with the Open RL Benchmark~\citep{Huang_Open_RL_Benchmark_2024}, enabling researchers to contrast them with various algorithms without reproducing the experiments.

We invite fellow researchers to use these resources to verify our findings or as a foundation for their investigative efforts.

\section*{Acknowledgements}\label{sec:acknowledgements}

This research was supported by the National Key Research and Development Program of China (No.2023YFB3308601), Science and Technology Service Network Initiative (No.KFJ-STS-QYZD-2021-21-001), the Talents by Sichuan provincial Party Committee Organization Department, and Chengdu - Chinese Academy of Sciences Science and Technology Cooperation Fund Project (Major Scientific and Technological Innovation Projects).

We would like to thank Shengyi Huang, Jiajing Cui, Wenxin Wu for useful discussions.

\bibliography{main}
\bibliographystyle{rlc}

\clearpage
\appendix

\section{Clarification on Terminology: Snapshot vs. Checkpoint and State}\label{sec:clarification-on-terminology-snapshot-vs-checkpoint-and-state}

\paragraph{Why do we use the term \textit{snapshot} instead of \textit{checkpoint}?}
We deliberately use the term \textit{snapshot} to distinguish it from the more commonly used \textit{checkpoint} in machine learning, emphasizing the stored models.
In the context of reinforcement learning, \textit{snapshot} is deliberately chosen to represent the comprehensive state of the interaction environment at a specific timestep, encompassing all aspects necessary to replicate an instance of the environment with precise fidelity fully.

\paragraph{What distinguishes a \textit{snapshot} from an RL environment state?}
A \textit{snapshot} captures a more comprehensive set of information than what is conveyed by the term \textit{state}. In addition to the observable environment state, a snapshot includes hidden variables present in scenarios like Partially Observable Markov Decision Processes (POMDP) and meta-information managed by environment wrappers.
This richer data collection ensures that the snapshot can reinitialize the environment, providing interactability that a simple state cannot.

\section{Experiment Details}\label{sec:experiment-details}

We used the CleanRL library's implementations for TD3, SAC, and PPO algorithms in our experiments~\cite{JMLR:v23:21-1342}. For PPO algorithm, however, we amended CleanRL's original implementation to rectify its incorrect truncation handling, informed by the approach used in Stable Baselines3~\cite{stable-baselines3}\footnote{The correction applied is detailed in Stable Baselines3's pull request 658: \url{https://github.com/DLR-RM/stable-baselines3/pull/658}}.

The implementations of S3RL+TD3, S3RL+SAC, and S3RL+PPO algorithms are all based on modifications of the previously described CleanRL implementations. Every implementation strictly adheres to CleanRL's single-file design philosophy to aid researchers in understanding and replicating our work.

All learning curve figures presented in this paper represent the average of evaluation results.
In each run, we conduct an evaluation every $5000$ timesteps, with each evaluation comprising three episodes.
We then calculate the average of these episodes to determine the evaluation result for that particular timestep.

\clearpage

\section{Additional Experiment}\label{sec:additional-experiment}

\subsection{Sweep of Teacher Models}\label{ssec:sweep-of-teacher-models}

In this subsection, we evaluated S3RL+TD3 under different performances of teacher agents.
We trained five teacher agents using CleanRL's TD3 implementation, each for 1 million timesteps on MuJoCo benchmark, with five different random seeds.
These teacher agents were subsequently ranked based on their evaluated performance, detailed in Table~\ref{tab:td3-models}.

\begin{figure}[ht]
    \centering
    \includegraphics[width=1.0\linewidth]{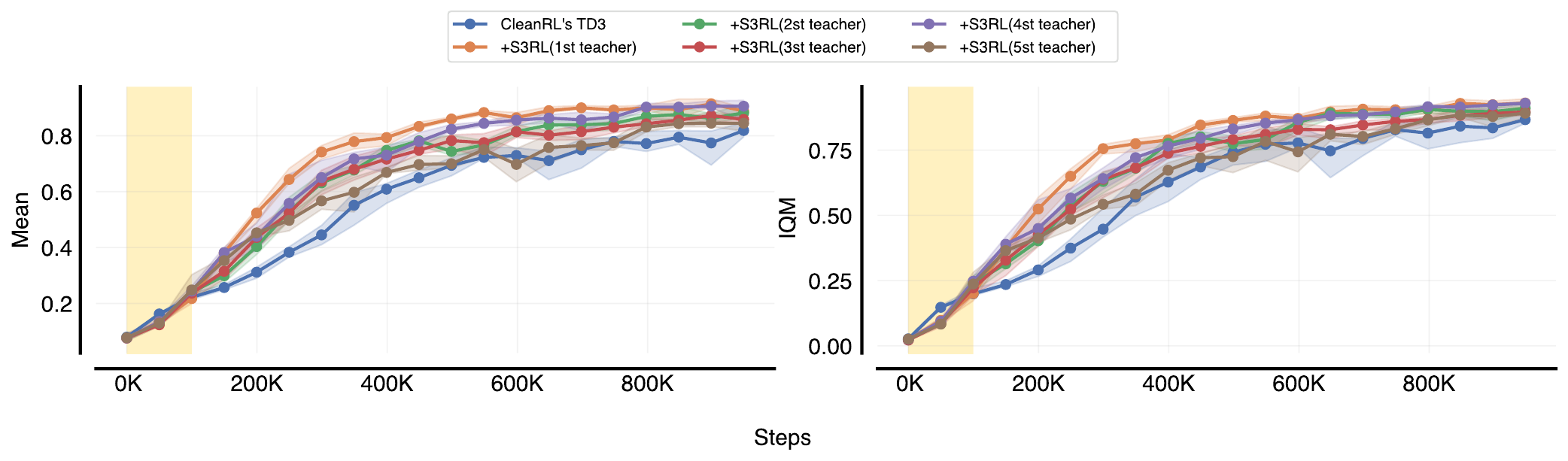}
    \caption{
        Learning curves sample efficiency sweeps for S3RL+TD3 across teacher models on six MuJoCo environments. For individual environment results, see Figure~\ref{fig:td3_sweep_teacher_indiv}.
    }\label{fig:td3_sweep_teacher}
\end{figure}

\begin{figure}[ht]
    \centering
    \includegraphics[width=1.0\linewidth]{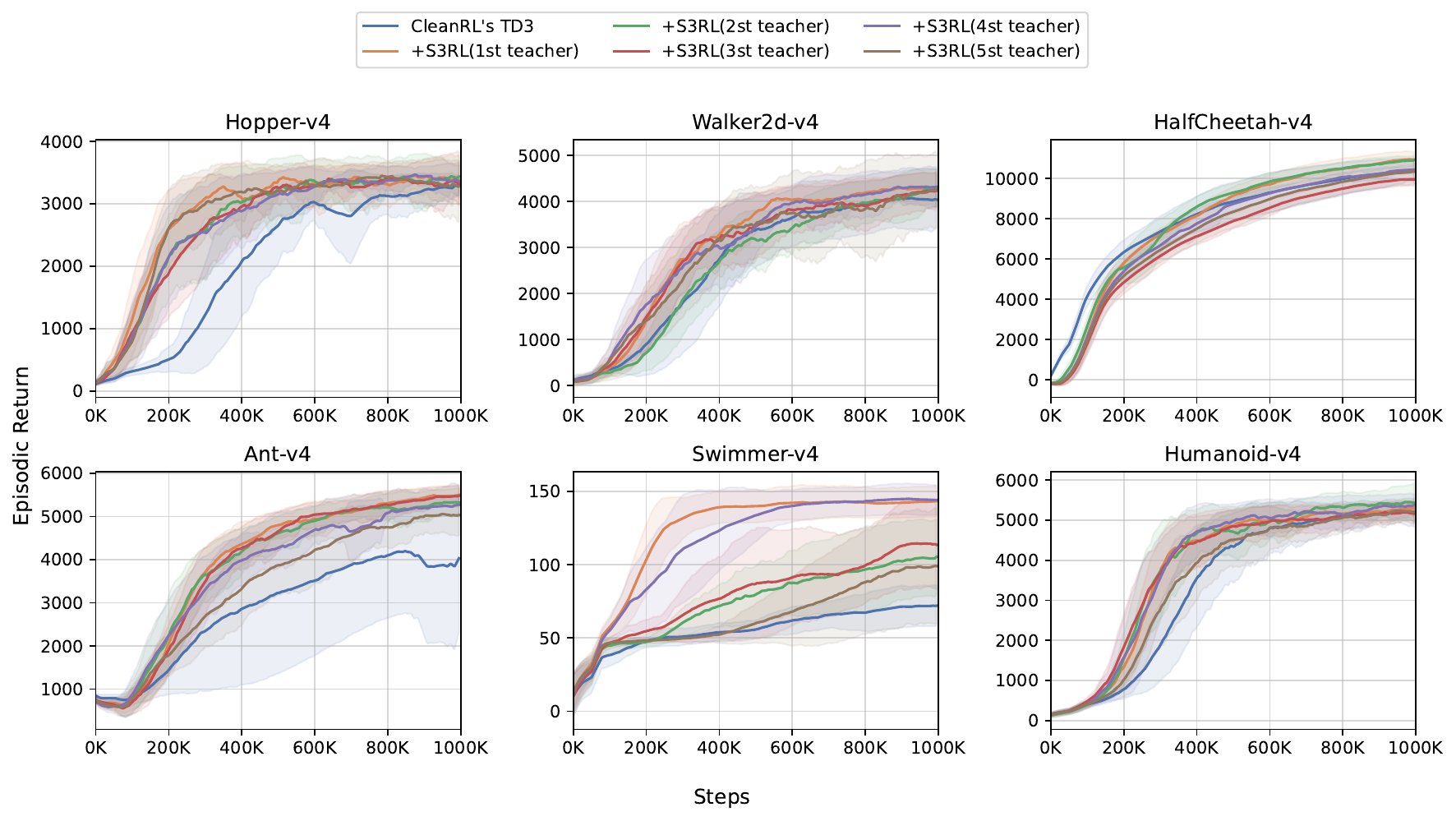}
    \caption{
        Learning curves sample efficiency sweeps for S3RL+TD3 across teacher models on each of six MuJoCo environments. Each subplot illustrates performance variance in sample efficiency across environments.
    }\label{fig:td3_sweep_teacher_indiv}
\end{figure}

We designed five sets of experiments, where each set uses teacher agents of different performance rankings to conduct the S3RL+TD3 experiment.
Our experimental results, as shown in Figures~\ref{fig:td3_sweep_teacher} and~\ref{fig:td3_sweep_teacher_indiv}, indicate that S3RL+TD3 shows variability in performance under the guidance of teacher agents with different levels of performance.
High-performing teacher agents generate a snapshot dataset that can lead to more significant performance improvements for the student agents, while those with relatively weaker performance offer more limited effects.
Notably, even teacher agents with performance below the average performance of TD3 can still enhance student agent's performance, suggesting that S3RL+TD3 can be effective even when high-quality teacher agents are unavailable.

\subsection{Evaluating S3RL with Soft Actor-Critic}\label{ssec:evaluating-s3rl-with-soft-actor-critic}

Soft Actor-Critic (SAC) is an advanced off-policy algorithm that optimizes a stochastic policy in an entropy-augmented RL framework, promoting a balance between exploration and exploitation.

\begin{figure}
    \centering
    \includegraphics[width=1.0\linewidth]{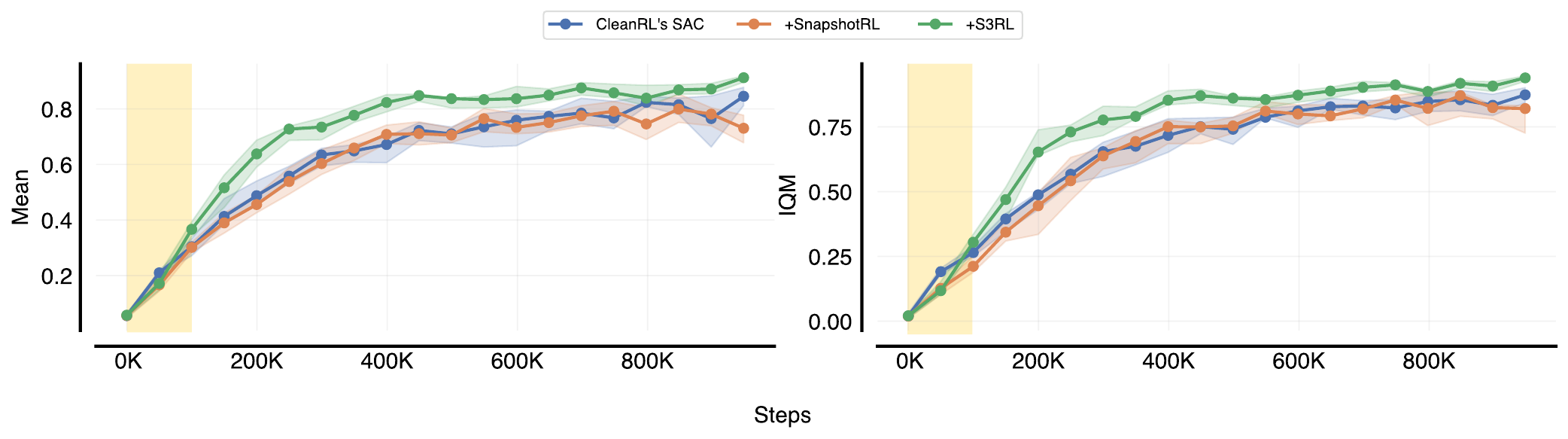}
    \caption{
        Learning curves sample efficiency comparison of SAC, \srl{}+SAC and S3RL+SAC on six MuJoCo environments. For individual environment results, see Figure~\ref{fig:sac_indiv}.
    }\label{fig:sac}
\end{figure}

\begin{figure}[ht]
    \centering
    \includegraphics[width=1.0\linewidth]{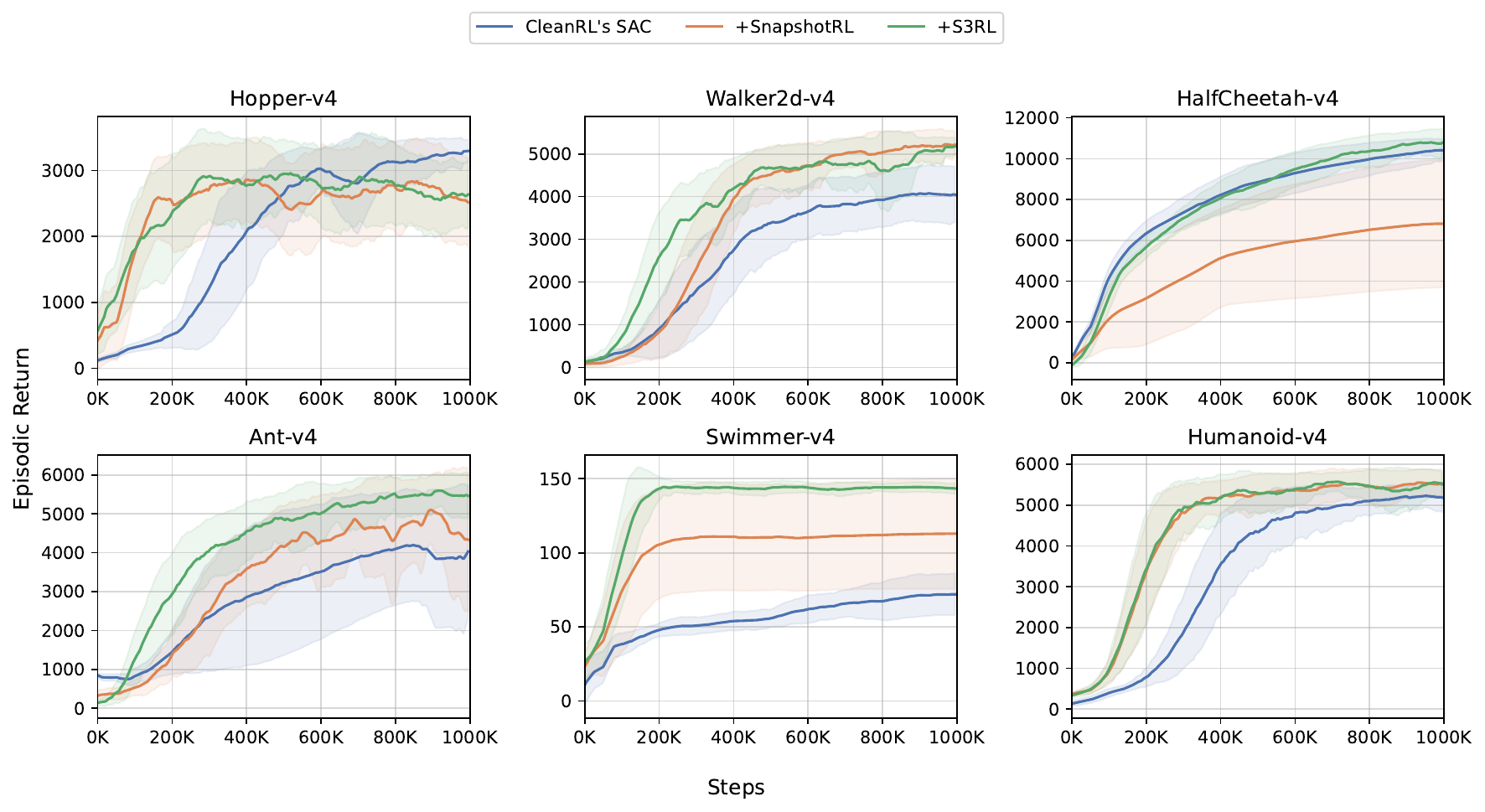}
    \caption{
        Detailed learning curves for SAC, \srl{}+SAC and S3RL+SAC on each of six MuJoCo environments. Each subplot illustrates performance variance in sample efficiency across environments.
    }\label{fig:sac_indiv}
\end{figure}

Similar to the experiments with TD3, we trained five teacher agents using CleanRL's SAC implementation, and selected the best performing teacher agent for generating the snapshot dataset.
SAC teacher models can be found in Table~\ref{tab:sac-models}.
Our results, as shown in Figure~\ref{fig:sac} and~\ref{fig:sac_indiv}, indicate that S3RL+SAC significantly outperforms SAC and \srl{}+SAC in terms of sample efficiency.

\subsection{Evaluating S3RL with Proximal Policy Optimization}\label{ssec:evaluating-s3rl-with-proximal-policy-optimization}

Proximal Policy Optimization (PPO) is a widely adopted on-policy algorithm that enhances learning stability and efficiency by employing a novel objective function with a clipping mechanism to prevent disruptive policy updates.

\begin{figure}
    \centering
    \includegraphics[width=1.0\linewidth]{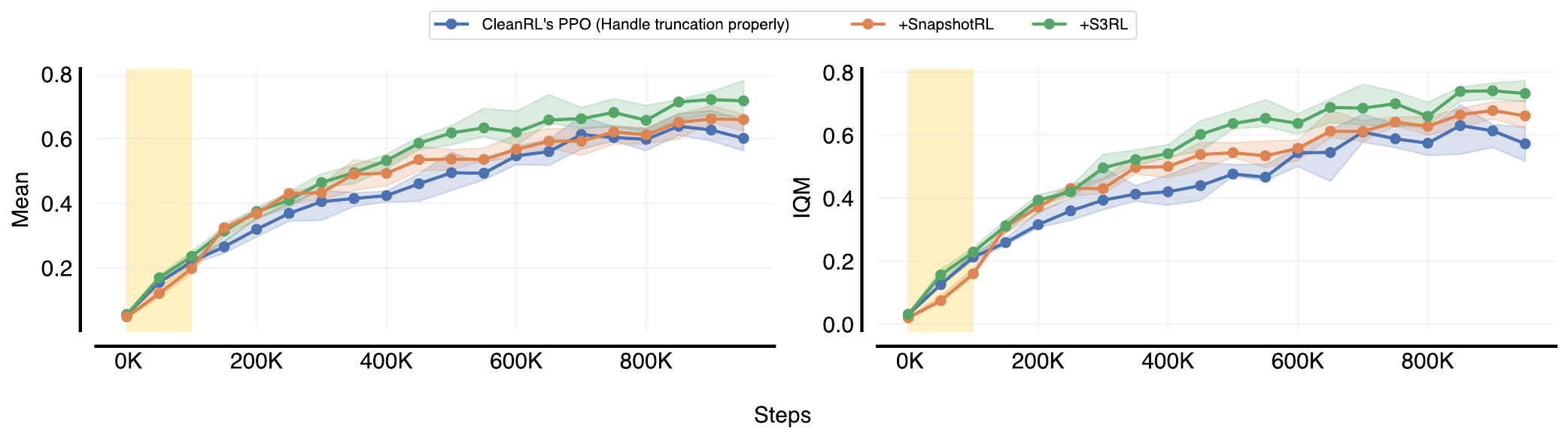}
    \caption{
        Learning curves sample efficiency comparison of PPO, \srl{}+PPO and S3RL+PPO on six MuJoCo environments. For individual environment results, see Figure~\ref{fig:ppo_indiv}.
    }\label{fig:ppo}
\end{figure}

\begin{figure}[ht]
    \centering
    \includegraphics[width=1.0\linewidth]{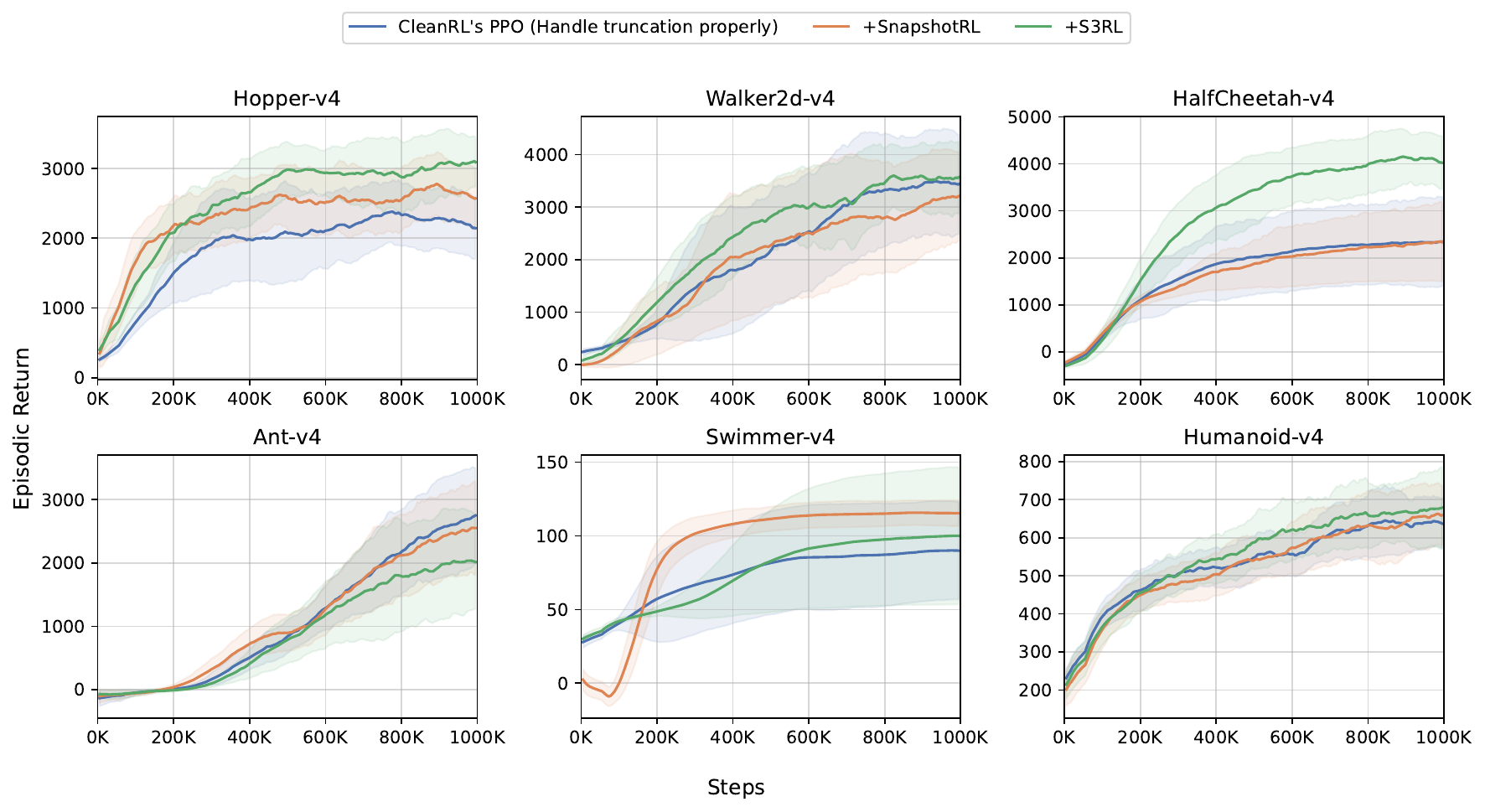}
    \caption{
        Detailed learning curves for PPO, \srl{}+PPO and S3RL+PPO on each of six MuJoCo environments. Each subplot illustrates performance variance in sample efficiency across environments.
    }\label{fig:ppo_indiv}
\end{figure}

Similar to the experiments with TD3, we trained five teacher agents using PPO algorithm, and selected the best performing teacher agent for generating the snapshot dataset.
PPO teacher models can be found in Table~\ref{tab:ppo-models}.
Our PPO is based on CleanRL's PPO implementation but presents a few implementation differences. For details, please refer to Appendix~\ref{sec:experiment-details}.
Our results, as shown in Figure~\ref{fig:ppo} and~\ref{fig:ppo_indiv}, indicate that S3RL+PPO only exhibits satisfactory performance in a limited set of environments.

The performance gains achieved by S3RL+PPO are small compared to S3RL+TD3 and S3RL+SAC, which we analyze for the following reasons:~\begin{itemize}
\item In S3RL+TD3 and S3RL+SAC experiments, due to their off-policy attributes, samples collected during the snapshotRL phase are stored in the replay buffer, thus exerting a continuous influence on subsequent learning phases. However, S3RL+PPO, being an on-policy algorithm, does not retain samples from the snapshotRL phase in the replay buffer, which consequently weakens their impact on future learning stages.
\item PPO employs Generalized Advantage Estimation (GAE), and the early handle truncation operation of STT strategy may affect the calculation of GAE.
\item PPO normalized observations and rewards, but training environment during \srl phase may alter the distribution of observations and rewards, affecting training after weaning off snapshots.
\item Across the MuJoCo benchmarks, the performance of PPO teacher agents is generally lower when compared with TD3 and SAC teacher agents.
\end{itemize}

\clearpage

\section{Additional Curves}\label{sec:additional-curves}

\begin{figure}[ht]
    \centering
    \includegraphics[width=1.0\linewidth]{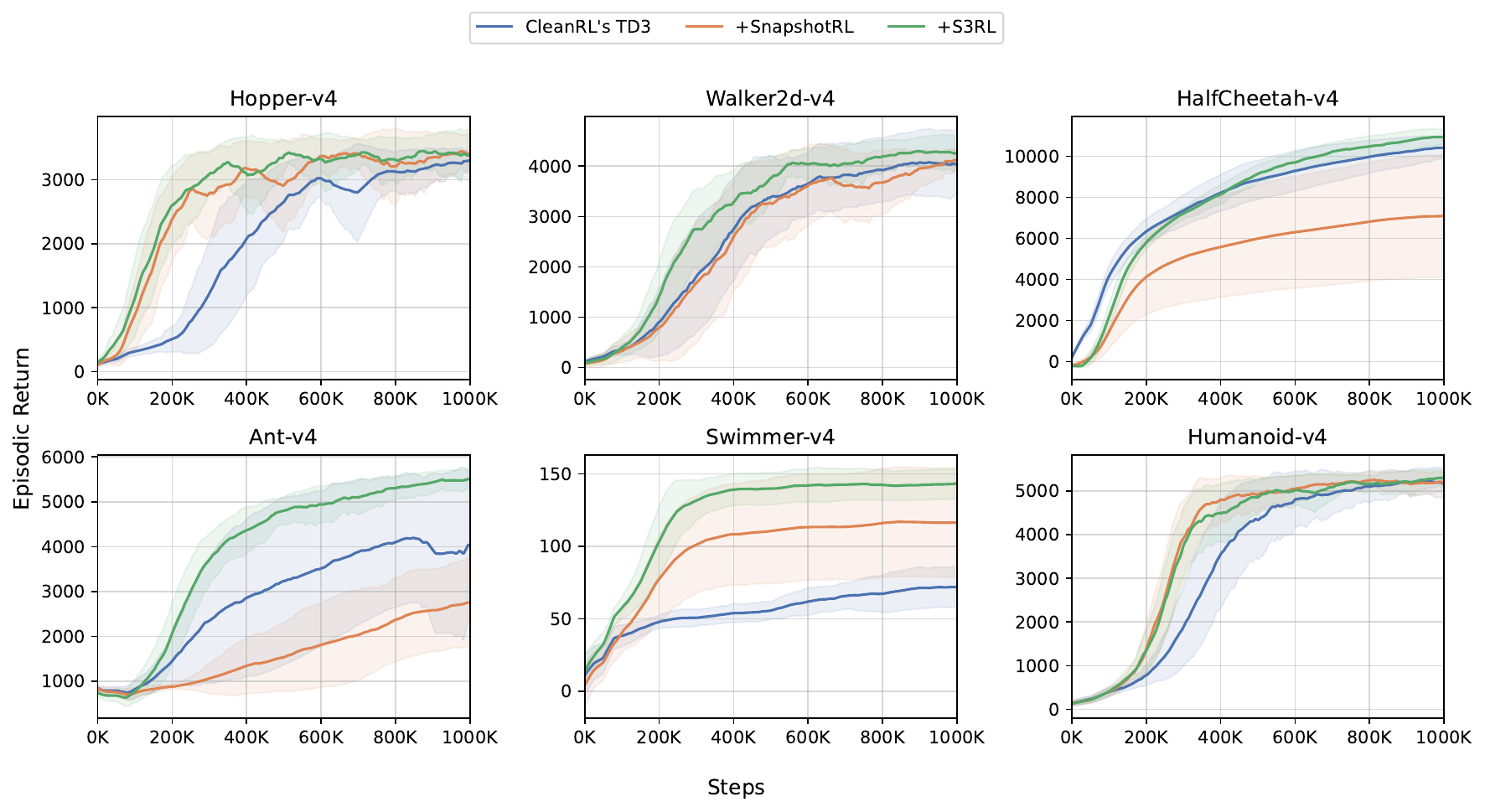}
    \caption{
        Detailed learning curves for TD3, \srl{}+TD3, and S3RL+TD3 on each of six MuJoCo environments. Each subplot illustrates performance variance in sample efficiency across environments.
    }\label{fig:td3_indiv}
\end{figure}

\begin{figure}[ht]
    \centering
    \includegraphics[width=1.0\linewidth]{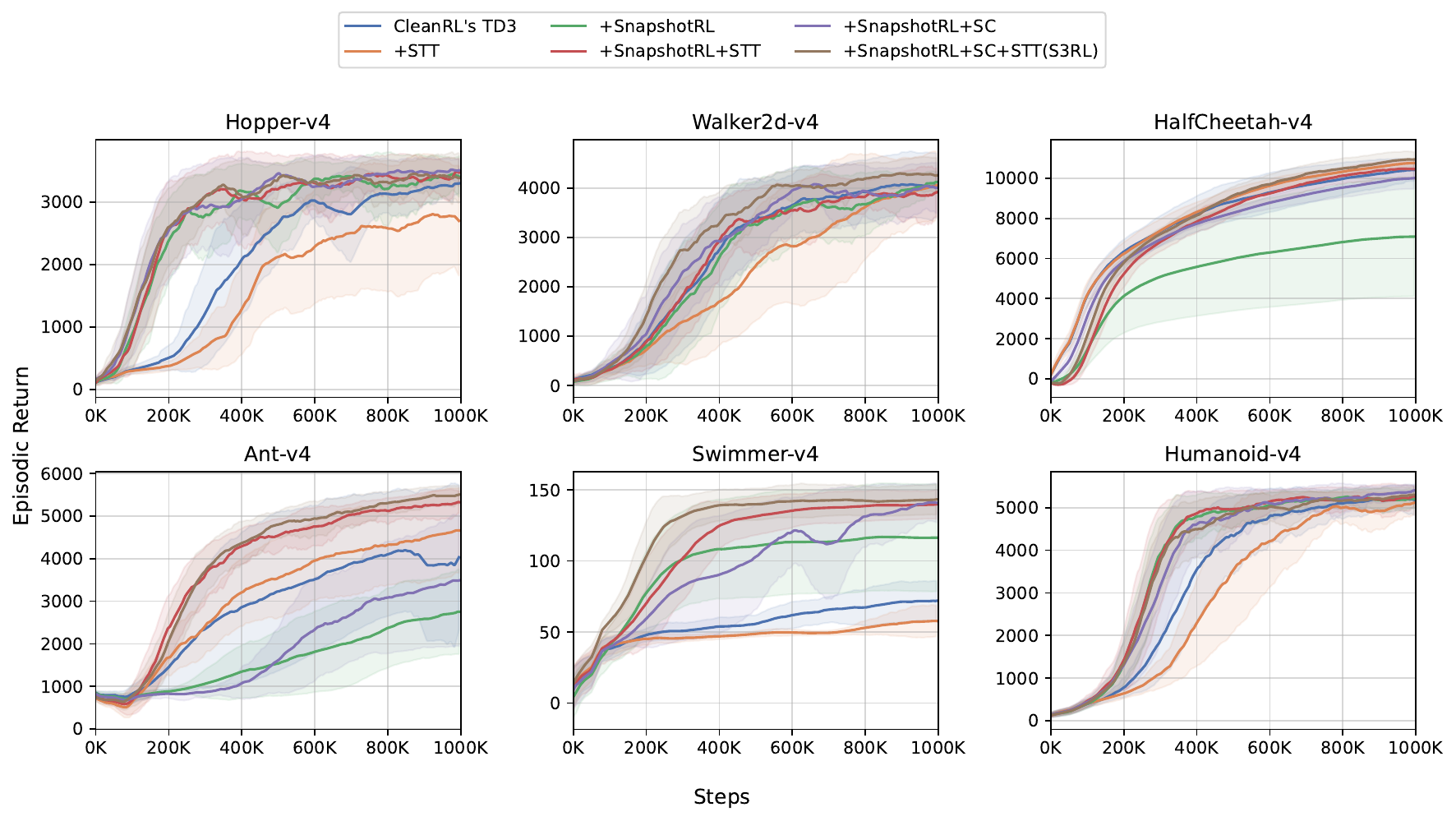}
    \caption{
        Ablation study details of S3RL+TD3, showing the impact of key components on sample efficiency on each of six MuJoCo environments. Each subplot illustrates performance variance in sample efficiency across environments.
    }\label{fig:td3_ablation_indiv}
\end{figure}

\begin{figure}[ht]
    \centering
    \includegraphics[width=1.0\linewidth]{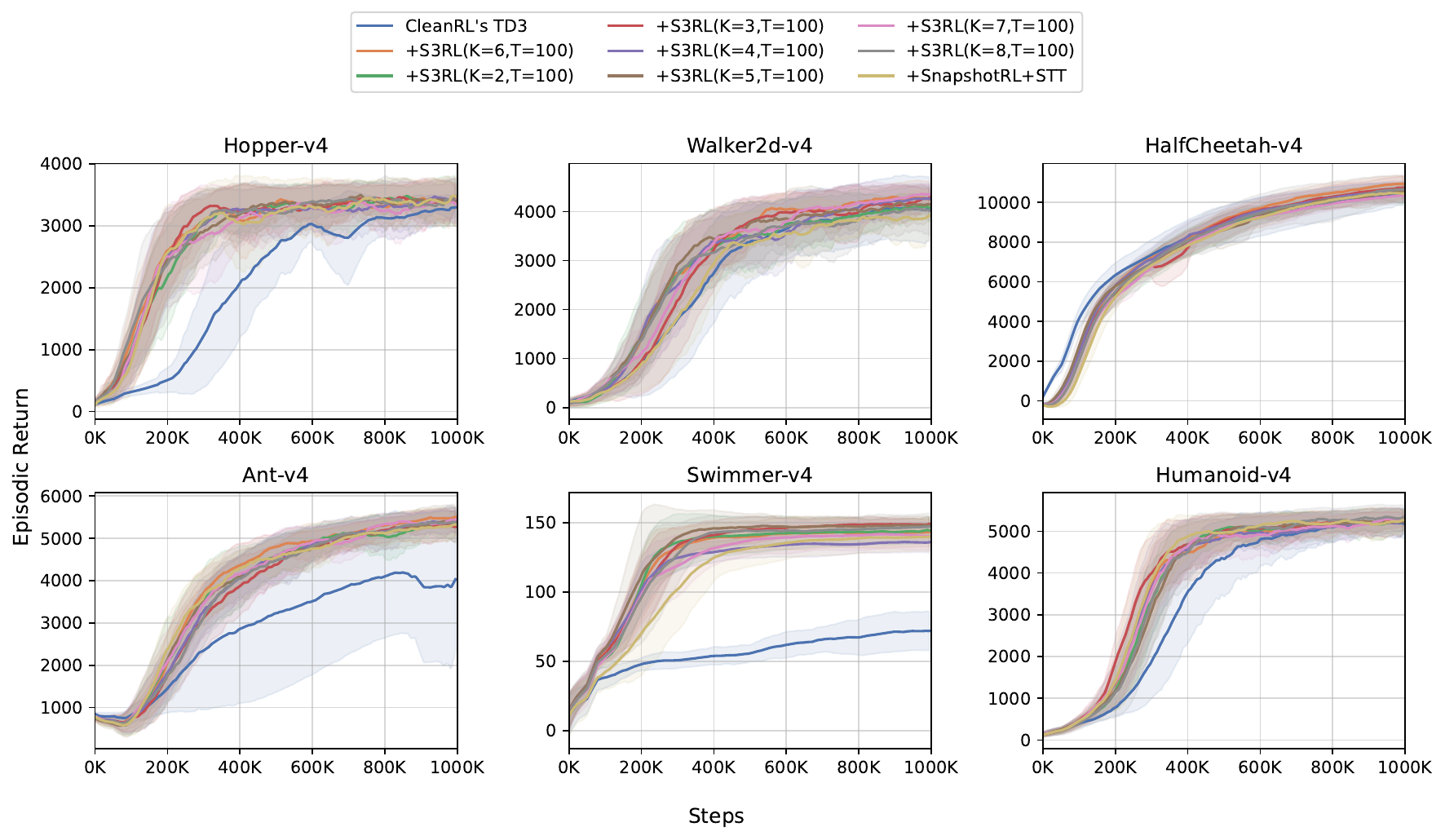}
    \caption{
        Learning curves sample efficiency sweeps for S3RL+TD3 across $K$ on each of six MuJoCo environments. Each subplot illustrates performance variance in sample efficiency across environments.
    }\label{fig:td3_sweep_k_indiv}
\end{figure}

\begin{figure}[ht]
    \centering
    \includegraphics[width=1.0\linewidth]{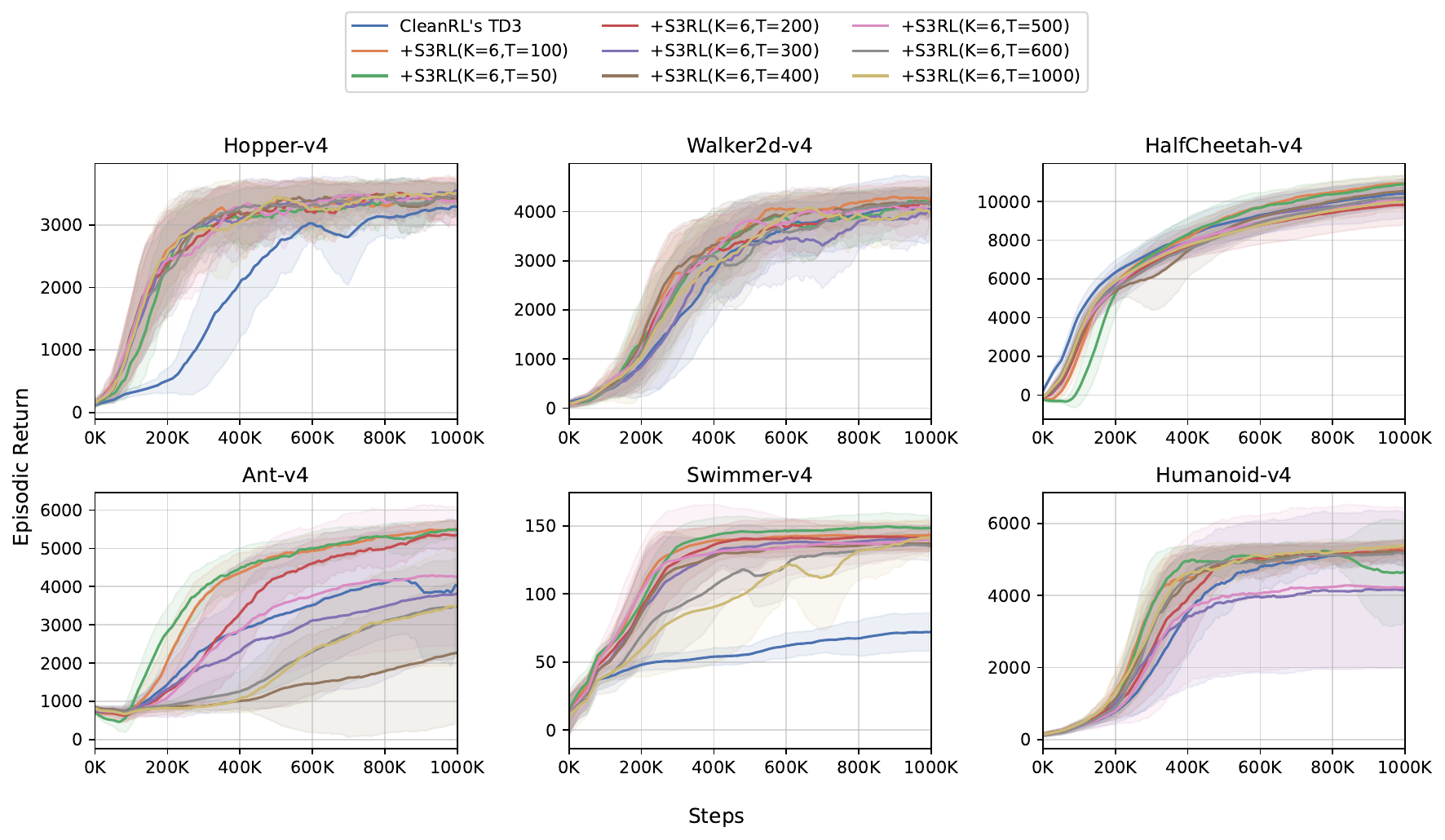}
    \caption{
        Learning curves sample efficiency sweeps for S3RL+TD3 across $T$ on each of six MuJoCo environments. Each subplot illustrates performance variance in sample efficiency across environments.
    }\label{fig:td3_sweep_t_indiv}
\end{figure}

\clearpage

\section{Hyperparameter}\label{sec:hyperparameter}

The hyperparameters for the implementations of S3RL+TD3, S3RL+SAC, and S3RL+PPO can be found in Table~\ref{tab:td3-params},~\ref{tab:sac-params}, and~\ref{tab:ppo-params}, respectively.
Each table is structured such that the standard hyperparameters correspond to the standard settings of each algorithm, while the latter parameters represent additional hyperparameters introduced for \colorbox{LightSalmon}{SnapshotRL} with \colorbox{LightAquamarine}{Status Classification} and \colorbox{LightOrchid}{Student Trajectory Truncation} (S3RL).

\begin{table}[ht]
    \centering
    \begin{tabular}{ll} 
    \toprule
    Parameter Names               & Parameter Values\\
    \midrule
    $N_{\text{total}}$ Total Time Steps       & 1,000,000  \\ 
    $\alpha$ Learning Rate                    & 0.0003 \\
    $N_{\text{buffer}}$ Replay Memory Buffer Size & 1,000,000  \\
    $\gamma$ Discount Factor                  & 0.99 \\ 
    $\tau$ Target Smoothing Coefficient       & 0.005 \\ 
    $N_{\text{batch}}$ Batch Size             & 256 \\
    Policy Noise Scale                        & 0.2 \\
    Exploration Noise Scale                   & 0.1 \\
    Time Steps Before Learning                & 25,000\\
    Training Policy Frequency (Delayed)       & 2\\
    Noise Clip Parameter for Target Policy Smoothing Regularization & 0.5\\
    Optimizer                                 & Adam\\
    \midrule
    $N_{\text{tep}}$ Number of Teacher Episodes          & \colorbox{LightSalmon}{10} \\
    $N_{\text{sostp}}$ Number of Steps in Snapshot Training Phase & \colorbox{LightSalmon}{100,000} \\ 
    $K$ for KMeans in State Classification           & \colorbox{LightAquamarine}{6} \\
    $T$ in Student Trajectory Truncation  & \colorbox{LightOrchid}{100} \\
    \bottomrule
    \end{tabular}
    \caption{TD3 and \srl{}+TD3 hyperparameters.}\label{tab:td3-params}
\end{table}

\begin{table}[ht]
    \centering
    \begin{tabular}{ll} 
    \toprule
    Parameter Names                                & Parameter Values\\
    \midrule
    $N_{\text{total}}$ Total Time Steps           & 1,000,000  \\ 
    $\alpha_{\text{policy}}$ Policy Network Learning Rate &  0.0003 \\
    $\alpha_{\text{Q}}$ Q Network Learning Rate   &  0.001 \\
    $N_{\text{buffer}}$ Replay Memory Buffer Size & 1,000,000  \\
    $\gamma$ Discount Factor                      & 0.99 \\ 
    $\tau$ Target Smoothing Coefficient           & 0.005 \\ 
    $N_{\text{batch}}$ Batch Size                 & 256 \\
    Policy Noise Scale                            & 0.2 \\
    Exploration Noise Scale                       & 0.1 \\
    Time Steps Before Learning                    & 5,000\\
    Policy Training Frequency (Delayed)           & 2\\
    Target Networks Update Frequency              & 1\\
    Noise Clip Parameter for Target Policy Smoothing Regularization & 0.5\\
    Entropy Regularization Coefficient            & 0.2\\
    Entropy Coefficient Auto-Tuning               & True\\
    Optimizer                                     & Adam\\
    \midrule
    $N_{\text{tep}}$ Number of Teacher Episodes          & \colorbox{LightSalmon}{10} \\
    $N_{\text{sostp}}$ Number of Steps in Snapshot Training Phase & \colorbox{LightSalmon}{100,000} \\ 
    $K$ for KMeans in State Classification           & \colorbox{LightAquamarine}{6} \\
    $T_{\text{stu}}$ in Student Trajectory Truncation  & \colorbox{LightOrchid}{100} \\
    \bottomrule
    \end{tabular}
    \caption{SAC and \srl{}+SAC hyperparameters.}\label{tab:sac-params}
\end{table}

\begin{table}[ht]
    \centering
    \begin{tabular}{ll} 
    \toprule
    Parameter Names  & Parameter Values\\
    \midrule
    $N_\text{total}$ Total Time Steps & 1,000,000  \\ 
    $\alpha$ Learning Rate &  0.0003 \\
    $N_\text{envs}$ Number of Parallel Environments & 1 \\
    $N_\text{steps}$ Number of Steps per Environment & 2048  \\
    $\gamma$ (Discount Factor) & 0.99 \\ 
    $\lambda$ (for GAE) & 0.95 \\ 
    $N_\text{mb}$ Number of Mini-batches & 32 \\
    $K$ (Number of PPO Update Iteration Per Epoch)& 10 \\
    $\varepsilon$ (PPO's Clipping Coefficient) & 0.2 \\
    $c_1$ (Value Function Coefficient) & 0.5\\
    $c_2$ (Entropy Coefficient) & 0.0\\
    $\omega$ (Gradient Norm Threshold)& 0.5 \\
    Value Function Loss Clipping & True\\
    Optimizer & Adam\\
    \midrule
    $N_{\text{tep}}$ Number of Teacher Episodes          & \colorbox{LightSalmon}{10} \\
    $N_{\text{sostp}}$ Number of Steps in Snapshot Training Phase & \colorbox{LightSalmon}{100,000} \\ 
    $K$ for KMeans in State Classification           & \colorbox{LightAquamarine}{6} \\
    $T_{\text{stu}}$ in Student Trajectory Truncation  & \colorbox{LightOrchid}{100} \\
    \bottomrule
    \end{tabular}
    \caption{PPO and \srl{}+PPO hyperparameters.}\label{tab:ppo-params}
\end{table}

\clearpage

\section{Teacher Models}\label{sec:teacher-models}

Table~\ref{tab:td3-models},~\ref{tab:sac-models}, and~\ref{tab:ppo-models} detail teacher models used in our experiments, ranked by their performance in terms of the mean evaluation score minus the standard deviation.
The model that ranks highest in each environment is indicated by a \includegraphics[height=0.8em]{media/emoji_pin.pdf} icon. Our main experimental analysis relies solely on these top-ranked models.

\begin{table}[ht]
    \centering
    \begin{tabular}{clrc}
    \toprule
        \footnotesize\textbf{Environment} & \footnotesize\textbf{Model Name (Click to go to repo)} & \footnotesize\textbf{Evaluation Score} & \footnotesize\textbf{Commit} \\
        \midrule
        \multirow{5}{*}{\footnotesize{Hopper-v4}} 
        & \includegraphics[height=0.8em]{media/emoji_pin.pdf}\blindhreftiny{https://huggingface.co/sdpkjc/Hopper-v4-td3_continuous_action-seed3}{\footnotesize\texttt{sdpkjc/Hopper-v4-td3\_continuous\_action-seed3}} & \footnotesize$3577.72 \pm 18.84$ & \blindhreftiny{https://huggingface.co/sdpkjc/Hopper-v4-td3_continuous_action-seed3/commit/77fccc4}{\footnotesize\texttt{77fccc4}} \\
        & \blindhreftiny{https://huggingface.co/cleanrl/Hopper-v4-td3_continuous_action-seed1}{\footnotesize\texttt{cleanrl/Hopper-v4-td3\_continuous\_action-seed1}} & \footnotesize$3244.59 \pm 8.55$ & \blindhreftiny{https://huggingface.co/cleanrl/Hopper-v4-td3_continuous_action-seed1/commit/1e14f8f}{\footnotesize\texttt{1e14f8f}} \\
        & \blindhreftiny{https://huggingface.co/sdpkjc/Hopper-v4-td3_continuous_action-seed2}{\footnotesize\texttt{sdpkjc/Hopper-v4-td3\_continuous\_action-seed2}} & \footnotesize$3162.70 \pm 400.28$ & \blindhreftiny{https://huggingface.co/sdpkjc/Hopper-v4-td3_continuous_action-seed2/commit/e3219bd}{\footnotesize\texttt{e3219bd}} \\
        & \blindhreftiny{https://huggingface.co/sdpkjc/Hopper-v4-td3_continuous_action-seed5}{\footnotesize\texttt{sdpkjc/Hopper-v4-td3\_continuous\_action-seed5}} & \footnotesize$3161.47 \pm 427.71$ & \blindhreftiny{https://huggingface.co/sdpkjc/Hopper-v4-td3_continuous_action-seed5/commit/754ff16}{\footnotesize\texttt{754ff16}} \\
        & \blindhreftiny{https://huggingface.co/sdpkjc/Hopper-v4-td3_continuous_action-seed4}{\footnotesize\texttt{sdpkjc/Hopper-v4-td3\_continuous\_action-seed4}} & \footnotesize$3094.02 \pm 807.61$ & \blindhreftiny{https://huggingface.co/sdpkjc/Hopper-v4-td3_continuous_action-seed4/commit/398b843}{\footnotesize\texttt{398b843}} \\
        \midrule

        \multirow{5}{*}{\footnotesize{Walker2d-v4}}
        & \includegraphics[height=0.8em]{media/emoji_pin.pdf}\blindhreftiny{https://huggingface.co/cleanrl/Walker2d-v4-td3_continuous_action-seed1}{\footnotesize\texttt{cleanrl/Walker2d-v4-td3\_continuous\_action-seed1}} & \footnotesize$3964.51 \pm 9.70$ & \blindhreftiny{https://huggingface.co/cleanrl/Walker2d-v4-td3_continuous_action-seed1/commit/51752b6}{\footnotesize\texttt{51752b6}} \\
        & \blindhreftiny{https://huggingface.co/sdpkjc/Walker2d-v4-td3_continuous_action-seed5}{\footnotesize\texttt{sdpkjc/Walker2d-v4-td3\_continuous\_action-seed5}} & \footnotesize$3678.97 \pm 340.29$ & \blindhreftiny{https://huggingface.co/sdpkjc/Walker2d-v4-td3_continuous_action-seed5/commit/089a235}{\footnotesize\texttt{089a235}} \\
        & \blindhreftiny{https://huggingface.co/sdpkjc/Walker2d-v4-td3_continuous_action-seed3}{\footnotesize\texttt{sdpkjc/Walker2d-v4-td3\_continuous\_action-seed3}} & \footnotesize$3314.10 \pm 12.34$ & \blindhreftiny{https://huggingface.co/sdpkjc/Walker2d-v4-td3_continuous_action-seed3/commit/614767a}{\footnotesize\texttt{614767a}} \\
        & \blindhreftiny{https://huggingface.co/sdpkjc/Walker2d-v4-td3_continuous_action-seed4}{\footnotesize\texttt{sdpkjc/Walker2d-v4-td3\_continuous\_action-seed4}} & \footnotesize$3624.09 \pm 539.63$ & \blindhreftiny{https://huggingface.co/sdpkjc/Walker2d-v4-td3_continuous_action-seed4/commit/dbf05cb}{\footnotesize\texttt{dbf05cb}} \\
        & \blindhreftiny{https://huggingface.co/sdpkjc/Walker2d-v4-td3_continuous_action-seed2}{\footnotesize\texttt{sdpkjc/Walker2d-v4-td3\_continuous\_action-seed2}} & \footnotesize$3527.67 \pm 746.96$ & \blindhreftiny{https://huggingface.co/sdpkjc/Walker2d-v4-td3_continuous_action-seed2/commit/fdf3439}{\footnotesize\texttt{fdf3439}} \\
        \midrule

        \multirow{5}{*}{\footnotesize{HalfCheetah-v4}}
        & \includegraphics[height=0.8em]{media/emoji_pin.pdf}\blindhreftiny{https://huggingface.co/cleanrl/HalfCheetah-v4-td3_continuous_action-seed1}{\footnotesize\texttt{cleanrl/HalfCheetah-v4-td3\_continuous\_action-seed1}} & \footnotesize$10762.42 \pm 84.09$ & \blindhreftiny{https://huggingface.co/cleanrl/HalfCheetah-v4-td3_continuous_action-seed1/commit/8547754}{\footnotesize\texttt{8547754}} \\
        & \blindhreftiny{https://huggingface.co/sdpkjc/HalfCheetah-v4-td3_continuous_action-seed4}{\footnotesize\texttt{sdpkjc/HalfCheetah-v4-td3\_continuous\_action-seed4}} & \footnotesize$10653.27 \pm 75.24$ & \blindhreftiny{https://huggingface.co/sdpkjc/HalfCheetah-v4-td3_continuous_action-seed4/commit/0f60f2f}{\footnotesize\texttt{0f60f2f}} \\
        & \blindhreftiny{https://huggingface.co/sdpkjc/HalfCheetah-v4-td3_continuous_action-seed3}{\footnotesize\texttt{sdpkjc/HalfCheetah-v4-td3\_continuous\_action-seed3}} & \footnotesize$11443.36 \pm 933.39$ & \blindhreftiny{https://huggingface.co/sdpkjc/HalfCheetah-v4-td3_continuous_action-seed3/commit/0c33876}{\footnotesize\texttt{0c33876}} \\
        & \blindhreftiny{https://huggingface.co/sdpkjc/HalfCheetah-v4-td3_continuous_action-seed2}{\footnotesize\texttt{sdpkjc/HalfCheetah-v4-td3\_continuous\_action-seed2}} & \footnotesize$10185.49 \pm 107.47$ & \blindhreftiny{https://huggingface.co/sdpkjc/HalfCheetah-v4-td3_continuous_action-seed2/commit/ec9624a}{\footnotesize\texttt{ec9624a}} \\
        & \blindhreftiny{https://huggingface.co/sdpkjc/HalfCheetah-v4-td3_continuous_action-seed5}{\footnotesize\texttt{sdpkjc/HalfCheetah-v4-td3\_continuous\_action-seed5}} & \footnotesize$10204.25 \pm 139.05$ & \blindhreftiny{https://huggingface.co/sdpkjc/HalfCheetah-v4-td3_continuous_action-seed5/commit/39939c8}{\footnotesize\texttt{39939c8}} \\
        \midrule

        \multirow{5}{*}{\footnotesize{Ant-v4}}
        & \includegraphics[height=0.8em]{media/emoji_pin.pdf}\blindhreftiny{https://huggingface.co/sdpkjc/Ant-v4-td3_continuous_action-seed4}{\footnotesize\texttt{sdpkjc/Ant-v4-td3\_continuous\_action-seed4}} & \footnotesize$5473.45 \pm 118.94$ & \blindhreftiny{https://huggingface.co/sdpkjc/Ant-v4-td3_continuous_action-seed4/commit/9a956a6}{\footnotesize\texttt{9a956a6}} \\
        & \blindhreftiny{https://huggingface.co/sdpkjc/Ant-v4-td3_continuous_action-seed3}{\footnotesize\texttt{sdpkjc/Ant-v4-td3\_continuous\_action-seed3}} & \footnotesize$5211.38 \pm 428.70$ & \blindhreftiny{https://huggingface.co/sdpkjc/Ant-v4-td3_continuous_action-seed3/commit/14610c5}{\footnotesize\texttt{14610c5}} \\
        & \blindhreftiny{https://huggingface.co/cleanrl/Ant-v4-td3_continuous_action-seed1}{\footnotesize\texttt{cleanrl/Ant-v4-td3\_continuous\_action-seed1}} & \footnotesize$5240.79 \pm 730.24$ & \blindhreftiny{https://huggingface.co/cleanrl/Ant-v4-td3_continuous_action-seed1/commit/3bd17bc}{\footnotesize\texttt{3bd17bc}} \\
        & \blindhreftiny{https://huggingface.co/sdpkjc/Ant-v4-td3_continuous_action-seed5}{\footnotesize\texttt{sdpkjc/Ant-v4-td3\_continuous\_action-seed5}} & \footnotesize$2802.61 \pm 163.65$ & \blindhreftiny{https://huggingface.co/sdpkjc/Ant-v4-td3_continuous_action-seed5/commit/074ff1a}{\footnotesize\texttt{074ff1a}} \\
        & \blindhreftiny{https://huggingface.co/sdpkjc/Ant-v4-td3_continuous_action-seed2}{\footnotesize\texttt{sdpkjc/Ant-v4-td3\_continuous\_action-seed2}} & \footnotesize$2606.88 \pm 36.88$ & \blindhreftiny{https://huggingface.co/sdpkjc/Ant-v4-td3_continuous_action-seed2/commit/ad845c9}{\footnotesize\texttt{ad845c9}} \\
        \midrule

        \multirow{5}{*}{\footnotesize{Swimmer-v4}}
        & \includegraphics[height=0.8em]{media/emoji_pin.pdf}\blindhreftiny{https://huggingface.co/sdpkjc/Swimmer-v4-td3_continuous_action-seed4}{\footnotesize\texttt{sdpkjc/Swimmer-v4-td3\_continuous\_action-seed4}} & \footnotesize$113.19 \pm 18.53$ & \blindhreftiny{https://huggingface.co/sdpkjc/Swimmer-v4-td3_continuous_action-seed4/commit/1161fa1}{\footnotesize\texttt{1161fa1}} \\
        & \blindhreftiny{https://huggingface.co/sdpkjc/Swimmer-v4-td3_continuous_action-seed2}{\footnotesize\texttt{sdpkjc/Swimmer-v4-td3\_continuous\_action-seed2}} & \footnotesize$88.97 \pm 19.63$ & \blindhreftiny{https://huggingface.co/sdpkjc/Swimmer-v4-td3_continuous_action-seed2/commit/d6ad4b1}{\footnotesize\texttt{d6ad4b1}} \\
        & \blindhreftiny{https://huggingface.co/sdpkjc/Swimmer-v4-td3_continuous_action-seed5}{\footnotesize\texttt{sdpkjc/Swimmer-v4-td3\_continuous\_action-seed5}} & \footnotesize$82.71 \pm 14.26$ & \blindhreftiny{https://huggingface.co/sdpkjc/Swimmer-v4-td3_continuous_action-seed5/commit/69dfa47}{\footnotesize\texttt{69dfa47}} \\
        & \blindhreftiny{https://huggingface.co/cleanrl/Swimmer-v4-td3_continuous_action-seed1}{\footnotesize\texttt{cleanrl/Swimmer-v4-td3\_continuous\_action-seed1}} & \footnotesize$60.09 \pm 9.06$ & \blindhreftiny{https://huggingface.co/cleanrl/Swimmer-v4-td3_continuous_action-seed1/commit/6eab7d2}{\footnotesize\texttt{6eab7d2}} \\
        & \blindhreftiny{https://huggingface.co/sdpkjc/Swimmer-v4-td3_continuous_action-seed3}{\footnotesize\texttt{sdpkjc/Swimmer-v4-td3\_continuous\_action-seed3}} & \footnotesize$62.38 \pm 12.78$ & \blindhreftiny{https://huggingface.co/sdpkjc/Swimmer-v4-td3_continuous_action-seed3/commit/380bcd0}{\footnotesize\texttt{380bcd0}} \\
        \midrule

        \multirow{5}{*}{\footnotesize{Humanoid-v4}}
        & \includegraphics[height=0.8em]{media/emoji_pin.pdf}\blindhreftiny{https://huggingface.co/sdpkjc/Humanoid-v4-td3_continuous_action-seed3}{\footnotesize\texttt{sdpkjc/Humanoid-v4-td3\_continuous\_action-seed3}} & \footnotesize$5279.53 \pm 35.43$ & \blindhreftiny{https://huggingface.co/sdpkjc/Humanoid-v4-td3_continuous_action-seed3/commit/e9dd75c}{\footnotesize\texttt{e9dd75c}} \\
        & \blindhreftiny{https://huggingface.co/sdpkjc/Humanoid-v4-td3_continuous_action-seed5}{\footnotesize\texttt{sdpkjc/Humanoid-v4-td3\_continuous\_action-seed5}} & \footnotesize$5189.38 \pm 27.99$ & \blindhreftiny{https://huggingface.co/sdpkjc/Humanoid-v4-td3_continuous_action-seed5/commit/c015a50}{\footnotesize\texttt{c015a50}} \\
        & \blindhreftiny{https://huggingface.co/sdpkjc/Humanoid-v4-td3_continuous_action-seed2}{\footnotesize\texttt{sdpkjc/Humanoid-v4-td3\_continuous\_action-seed2}} & \footnotesize$5038.18 \pm 130.45$ & \blindhreftiny{https://huggingface.co/sdpkjc/Humanoid-v4-td3_continuous_action-seed2/commit/5f196ad}{\footnotesize\texttt{5f196ad}} \\
        & \blindhreftiny{https://huggingface.co/cleanrl/Humanoid-v4-td3_continuous_action-seed1}{\footnotesize\texttt{cleanrl/Humanoid-v4-td3\_continuous\_action-seed1}} & \footnotesize$5303.39 \pm 514.14$ & \blindhreftiny{https://huggingface.co/cleanrl/Humanoid-v4-td3_continuous_action-seed1/commit/0450bee}{\footnotesize\texttt{0450bee}} \\
        & \blindhreftiny{https://huggingface.co/sdpkjc/Humanoid-v4-td3_continuous_action-seed4}{\footnotesize\texttt{sdpkjc/Humanoid-v4-td3\_continuous\_action-seed4}} & \footnotesize$4880.24 \pm 1187.43$ & \blindhreftiny{https://huggingface.co/sdpkjc/Humanoid-v4-td3_continuous_action-seed4/commit/873f6ab}{\footnotesize\texttt{873f6ab}} \\         
        \bottomrule
    \end{tabular}
    \caption{TD3 Models Evaluation Scores and Links}\label{tab:td3-models}
\end{table}

\begin{table}[ht]
    \centering
    \begin{tabular}{clrc}
    \toprule
        \footnotesize\textbf{Environment} & \footnotesize\textbf{Model Name (Click to go to repo)} & \footnotesize\textbf{Evaluation Score} & \footnotesize\textbf{Commit} \\
        \midrule
        \multirow{5}{*}{\footnotesize{Hopper-v4}} 
        & \includegraphics[height=0.8em]{media/emoji_pin.pdf}\blindhreftiny{https://huggingface.co/sdpkjc/Hopper-v4-sac_continuous_action-seed4}{\footnotesize\texttt{sdpkjc/Hopper-v4-sac\_continuous\_action-seed4}} & \footnotesize$2862.20 \pm 972.12$ & \blindhreftiny{https://huggingface.co/sdpkjc/Hopper-v4-sac_continuous_action-seed4/commit/1ee692e}{\footnotesize\texttt{1ee692e}} \\
        & \blindhreftiny{https://huggingface.co/sdpkjc/Hopper-v4-sac_continuous_action-seed3}{\footnotesize\texttt{sdpkjc/Hopper-v4-sac\_continuous\_action-seed3}} & \footnotesize$2493.92 \pm 609.06$ & \blindhreftiny{https://huggingface.co/sdpkjc/Hopper-v4-sac_continuous_action-seed3/commit/4015a94}{\footnotesize\texttt{4015a94}} \\
        & \blindhreftiny{https://huggingface.co/sdpkjc/Hopper-v4-sac_continuous_action-seed1}{\footnotesize\texttt{sdpkjc/Hopper-v4-sac\_continuous\_action-seed1}} & \footnotesize$2274.04 \pm 605.18$ & \blindhreftiny{https://huggingface.co/sdpkjc/Hopper-v4-sac_continuous_action-seed1/commit/63ba003}{\footnotesize\texttt{63ba003}} \\
        & \blindhreftiny{https://huggingface.co/sdpkjc/Hopper-v4-sac_continuous_action-seed5}{\footnotesize\texttt{sdpkjc/Hopper-v4-sac\_continuous\_action-seed5}} & \footnotesize$1598.77 \pm 492.69$ & \blindhreftiny{https://huggingface.co/sdpkjc/Hopper-v4-sac_continuous_action-seed5/commit/bf93082}{\footnotesize\texttt{bf93082}} \\
        & \blindhreftiny{https://huggingface.co/sdpkjc/Hopper-v4-sac_continuous_action-seed2}{\footnotesize\texttt{sdpkjc/Hopper-v4-sac\_continuous\_action-seed2}} & \footnotesize$1555.12 \pm 279.93$ & \blindhreftiny{https://huggingface.co/sdpkjc/Hopper-v4-sac_continuous_action-seed2/commit/d6b664e}{\footnotesize\texttt{d6b664e}} \\
        \midrule

        \multirow{5}{*}{\footnotesize{Walker2d-v4}}
        & \includegraphics[height=0.8em]{media/emoji_pin.pdf}\blindhreftiny{https://huggingface.co/sdpkjc/Walker2d-v4-sac_continuous_action-seed4}{\footnotesize\texttt{sdpkjc/Walker2d-v4-sac\_continuous\_action-seed4}} & \footnotesize$5350.98 \pm 89.84$ & \blindhreftiny{https://huggingface.co/sdpkjc/Walker2d-v4-sac_continuous_action-seed4/commit/c8561ff}{\footnotesize\texttt{c8561ff}} \\
        & \blindhreftiny{https://huggingface.co/sdpkjc/Walker2d-v4-sac_continuous_action-seed3}{\footnotesize\texttt{sdpkjc/Walker2d-v4-sac\_continuous\_action-seed3}} & \footnotesize$5237.31 \pm 942.48$ & \blindhreftiny{https://huggingface.co/sdpkjc/Walker2d-v4-sac_continuous_action-seed3/commit/2bb35b1}{\footnotesize\texttt{2bb35b1}} \\
        & \blindhreftiny{https://huggingface.co/sdpkjc/Walker2d-v4-sac_continuous_action-seed1}{\footnotesize\texttt{sdpkjc/Walker2d-v4-sac\_continuous\_action-seed1}} & \footnotesize$5192.85 \pm 85.73$ & \blindhreftiny{https://huggingface.co/sdpkjc/Walker2d-v4-sac_continuous_action-seed1/commit/29d35a9}{\footnotesize\texttt{29d35a9}} \\
        & \blindhreftiny{https://huggingface.co/sdpkjc/Walker2d-v4-sac_continuous_action-seed5}{\footnotesize\texttt{sdpkjc/Walker2d-v4-sac\_continuous\_action-seed5}} & \footnotesize$4731.36 \pm 28.52$ & \blindhreftiny{https://huggingface.co/sdpkjc/Walker2d-v4-sac_continuous_action-seed5/commit/7a7f631}{\footnotesize\texttt{7a7f631}} \\
        & \blindhreftiny{https://huggingface.co/sdpkjc/Walker2d-v4-sac_continuous_action-seed2}{\footnotesize\texttt{sdpkjc/Walker2d-v4-sac\_continuous\_action-seed2}} & \footnotesize$3678.91 \pm 523.03$ & \blindhreftiny{https://huggingface.co/sdpkjc/Walker2d-v4-sac_continuous_action-seed2/commit/49501ed}{\footnotesize\texttt{49501ed}} \\
        \midrule

        \multirow{5}{*}{\footnotesize{HalfCheetah-v4}}
        & \includegraphics[height=0.8em]{media/emoji_pin.pdf}\blindhreftiny{https://huggingface.co/sdpkjc/HalfCheetah-v4-sac_continuous_action-seed4}{\footnotesize\texttt{sdpkjc/HalfCheetah-v4-sac\_continuous\_action-seed4}} & \footnotesize$11623.83 \pm 156.02$ & \blindhreftiny{https://huggingface.co/sdpkjc/HalfCheetah-v4-sac_continuous_action-seed4/commit/bf0622e}{\footnotesize\texttt{bf0622e}} \\
        & \blindhreftiny{https://huggingface.co/sdpkjc/HalfCheetah-v4-sac_continuous_action-seed2}{\footnotesize\texttt{sdpkjc/HalfCheetah-v4-sac\_continuous\_action-seed2}} & \footnotesize$11615.36 \pm 1484.63$ & \blindhreftiny{https://huggingface.co/sdpkjc/HalfCheetah-v4-sac_continuous_action-seed2/commit/f5122c3}{\footnotesize\texttt{f5122c3}} \\
        & \blindhreftiny{https://huggingface.co/sdpkjc/HalfCheetah-v4-sac_continuous_action-seed3}{\footnotesize\texttt{sdpkjc/HalfCheetah-v4-sac\_continuous\_action-seed3}} & \footnotesize$11543.00 \pm 122.49$ & \blindhreftiny{https://huggingface.co/sdpkjc/HalfCheetah-v4-sac_continuous_action-seed3/commit/a8c2810}{\footnotesize\texttt{a8c2810}} \\
        & \blindhreftiny{https://huggingface.co/sdpkjc/HalfCheetah-v4-sac_continuous_action-seed1}{\footnotesize\texttt{sdpkjc/HalfCheetah-v4-sac\_continuous\_action-seed1}} & \footnotesize$11211.47 \pm 972.19$ & \blindhreftiny{https://huggingface.co/sdpkjc/HalfCheetah-v4-sac_continuous_action-seed1/commit/19da4f5}{\footnotesize\texttt{19da4f5}} \\
        & \blindhreftiny{https://huggingface.co/sdpkjc/HalfCheetah-v4-sac_continuous_action-seed5}{\footnotesize\texttt{sdpkjc/HalfCheetah-v4-sac\_continuous\_action-seed5}} & \footnotesize$8187.18 \pm 676.54$ & \blindhreftiny{https://huggingface.co/sdpkjc/HalfCheetah-v4-sac_continuous_action-seed5/commit/6816d88}{\footnotesize\texttt{6816d88}} \\
        \midrule
        
        \multirow{5}{*}{\footnotesize{Ant-v4}}
        & \includegraphics[height=0.8em]{media/emoji_pin.pdf}\blindhreftiny{https://huggingface.co/sdpkjc/Ant-v4-sac_continuous_action-seed3}{\footnotesize\texttt{sdpkjc/Ant-v4-sac\_continuous\_action-seed3}} & \footnotesize$5735.30 \pm 989.07$ & \blindhreftiny{https://huggingface.co/sdpkjc/Ant-v4-sac_continuous_action-seed3/commit/b1126bf}{\footnotesize\texttt{b1126bf}} \\
        & \blindhreftiny{https://huggingface.co/sdpkjc/Ant-v4-sac_continuous_action-seed4}{\footnotesize\texttt{sdpkjc/Ant-v4-sac\_continuous\_action-seed4}} & \footnotesize$5517.12 \pm 1143.23$ & \blindhreftiny{https://huggingface.co/sdpkjc/Ant-v4-sac_continuous_action-seed4/commit/83b4537}{\footnotesize\texttt{83b4537}} \\
        & \blindhreftiny{https://huggingface.co/sdpkjc/Ant-v4-sac_continuous_action-seed2}{\footnotesize\texttt{sdpkjc/Ant-v4-sac\_continuous\_action-seed2}} & \footnotesize$5511.89 \pm 1041.57$ & \blindhreftiny{https://huggingface.co/sdpkjc/Ant-v4-sac_continuous_action-seed2/commit/514f6d2}{\footnotesize\texttt{514f6d2}} \\
        & \blindhreftiny{https://huggingface.co/sdpkjc/Ant-v4-sac_continuous_action-seed1}{\footnotesize\texttt{sdpkjc/Ant-v4-sac\_continuous\_action-seed1}} & \footnotesize$5314.44 \pm 1159.54$ & \blindhreftiny{https://huggingface.co/sdpkjc/Ant-v4-sac_continuous_action-seed1/commit/b32f853}{\footnotesize\texttt{b32f853}} \\
        & \blindhreftiny{https://huggingface.co/sdpkjc/Ant-v4-sac_continuous_action-seed5}{\footnotesize\texttt{sdpkjc/Ant-v4-sac\_continuous\_action-seed5}} & \footnotesize$3544.68 \pm 2044.81$ & \blindhreftiny{https://huggingface.co/sdpkjc/Ant-v4-sac_continuous_action-seed5/commit/be8c365}{\footnotesize\texttt{be8c365}} \\
        \midrule
        
        \multirow{5}{*}{\footnotesize{Swimmer-v4}}
        & \includegraphics[height=0.8em]{media/emoji_pin.pdf}\blindhreftiny{https://huggingface.co/sdpkjc/Swimmer-v4-sac_continuous_action-seed3}{\footnotesize\texttt{sdpkjc/Swimmer-v4-sac\_continuous\_action-seed3}} & \footnotesize$148.97 \pm 5.85$ & \blindhreftiny{https://huggingface.co/sdpkjc/Swimmer-v4-sac_continuous_action-seed3/commit/6c0875a}{\footnotesize\texttt{6c0875a}} \\
        & \blindhreftiny{https://huggingface.co/sdpkjc/Swimmer-v4-sac_continuous_action-seed2}{\footnotesize\texttt{sdpkjc/Swimmer-v4-sac\_continuous\_action-seed2}} & \footnotesize$76.70 \pm 25.53$ & \blindhreftiny{https://huggingface.co/sdpkjc/Swimmer-v4-sac_continuous_action-seed2/commit/cf113b4}{\footnotesize\texttt{cf113b4}} \\
        & \blindhreftiny{https://huggingface.co/sdpkjc/Swimmer-v4-sac_continuous_action-seed1}{\footnotesize\texttt{sdpkjc/Swimmer-v4-sac\_continuous\_action-seed1}} & \footnotesize$74.85 \pm 27.64$ & \blindhreftiny{https://huggingface.co/sdpkjc/Swimmer-v4-sac_continuous_action-seed1/commit/d9fd594}{\footnotesize\texttt{d9fd594}} \\
        & \blindhreftiny{https://huggingface.co/sdpkjc/Swimmer-v4-sac_continuous_action-seed4}{\footnotesize\texttt{sdpkjc/Swimmer-v4-sac\_continuous\_action-seed4}} & \footnotesize$50.26 \pm 2.03$ & \blindhreftiny{https://huggingface.co/sdpkjc/Swimmer-v4-sac_continuous_action-seed4/commit/40ca421}{\footnotesize\texttt{40ca421}} \\
        & \blindhreftiny{https://huggingface.co/sdpkjc/Swimmer-v4-sac_continuous_action-seed5}{\footnotesize\texttt{sdpkjc/Swimmer-v4-sac\_continuous\_action-seed5}} & \footnotesize$46.46 \pm 1.08$ & \blindhreftiny{https://huggingface.co/sdpkjc/Swimmer-v4-sac_continuous_action-seed5/commit/94560c4}{\footnotesize\texttt{94560c4}} \\
        \midrule

        \multirow{5}{*}{\footnotesize{Humanoid-v4}}
        & \includegraphics[height=0.8em]{media/emoji_pin.pdf}\blindhreftiny{https://huggingface.co/sdpkjc/Humanoid-v4-sac_continuous_action-seed4}{\footnotesize\texttt{sdpkjc/Humanoid-v4-sac\_continuous\_action-seed4}} & \footnotesize$5604.16 \pm 404.34$ & \blindhreftiny{https://huggingface.co/sdpkjc/Humanoid-v4-sac_continuous_action-seed4/commit/316b06c}{\footnotesize\texttt{316b06c}} \\
        & \blindhreftiny{https://huggingface.co/sdpkjc/Humanoid-v4-sac_continuous_action-seed5}{\footnotesize\texttt{sdpkjc/Humanoid-v4-sac\_continuous\_action-seed5}} & \footnotesize$5570.79 \pm 750.60$ & \blindhreftiny{https://huggingface.co/sdpkjc/Humanoid-v4-sac_continuous_action-seed5/commit/6e3b960}{\footnotesize\texttt{6e3b960}} \\
        & \blindhreftiny{https://huggingface.co/sdpkjc/Humanoid-v4-sac_continuous_action-seed3}{\footnotesize\texttt{sdpkjc/Humanoid-v4-sac\_continuous\_action-seed3}} & \footnotesize$5328.96 \pm 1015.76$ & \blindhreftiny{https://huggingface.co/sdpkjc/Humanoid-v4-sac_continuous_action-seed3/commit/204ee92}{\footnotesize\texttt{204ee92}} \\
        & \blindhreftiny{https://huggingface.co/sdpkjc/Humanoid-v4-sac_continuous_action-seed2}{\footnotesize\texttt{sdpkjc/Humanoid-v4-sac\_continuous\_action-seed2}} & \footnotesize$5306.36 \pm 466.78$ & \blindhreftiny{https://huggingface.co/sdpkjc/Humanoid-v4-sac_continuous_action-seed2/commit/72f53bc}{\footnotesize\texttt{72f53bc}} \\
        & \blindhreftiny{https://huggingface.co/sdpkjc/Humanoid-v4-sac_continuous_action-seed1}{\footnotesize\texttt{sdpkjc/Humanoid-v4-sac\_continuous\_action-seed1}} & \footnotesize$5220.03 \pm 212.43$ & \blindhreftiny{https://huggingface.co/sdpkjc/Humanoid-v4-sac_continuous_action-seed1/commit/6f2042f}{\footnotesize\texttt{6f2042f}} \\
        \bottomrule
    \end{tabular}
    \caption{SAC Models Evaluation Scores and Links}\label{tab:sac-models}
\end{table}

\begin{table}[ht]
    \centering
    \begin{tabular}{clrc}
    \toprule
        \footnotesize\textbf{Environment} & \footnotesize\textbf{Model Name (Click to go to repo)} & \footnotesize\textbf{Evaluation Score} & \footnotesize\textbf{Commit} \\
        \midrule
        \multirow{5}{*}{\footnotesize{Hopper-v4}}
        & \includegraphics[height=0.8em]{media/emoji_pin.pdf}\blindhreftiny{https://huggingface.co/sdpkjc/Hopper-v4-ppo_fix_continuous_action-seed3}{\footnotesize\texttt{Hopper-v4-ppo\_fix\_continuous\_action-seed3}} & \footnotesize$2515.99 \pm 807.22$ & \blindhreftiny{https://huggingface.co/sdpkjc/Hopper-v4-ppo_fix_continuous_action-seed3/commit/3d317e2}{\footnotesize\texttt{3d317e2}} \\
        & \blindhreftiny{https://huggingface.co/sdpkjc/Hopper-v4-ppo_fix_continuous_action-seed5}{\footnotesize\texttt{Hopper-v4-ppo\_fix\_continuous\_action-seed5}} & \footnotesize$2444.71 \pm 794.51$ & \blindhreftiny{https://huggingface.co/sdpkjc/Hopper-v4-ppo_fix_continuous_action-seed5/commit/3f3fd61}{\footnotesize\texttt{3f3fd61}} \\
        & \blindhreftiny{https://huggingface.co/sdpkjc/Hopper-v4-ppo_fix_continuous_action-seed2}{\footnotesize\texttt{Hopper-v4-ppo\_fix\_continuous\_action-seed2}} & \footnotesize$1990.14 \pm 683.73$ & \blindhreftiny{https://huggingface.co/sdpkjc/Hopper-v4-ppo_fix_continuous_action-seed2/commit/54a25d8}{\footnotesize\texttt{54a25d8}} \\
        & \blindhreftiny{https://huggingface.co/sdpkjc/Hopper-v4-ppo_fix_continuous_action-seed4}{\footnotesize\texttt{Hopper-v4-ppo\_fix\_continuous\_action-seed4}} & \footnotesize$1917.18 \pm 681.46$ & \blindhreftiny{https://huggingface.co/sdpkjc/Hopper-v4-ppo_fix_continuous_action-seed4/commit/2322d58}{\footnotesize\texttt{2322d58}} \\
        & \blindhreftiny{https://huggingface.co/sdpkjc/Hopper-v4-ppo_fix_continuous_action-seed1}{\footnotesize\texttt{Hopper-v4-ppo\_fix\_continuous\_action-seed1}} & \footnotesize$1649.65 \pm 559.09$ & \blindhreftiny{https://huggingface.co/sdpkjc/Hopper-v4-ppo_fix_continuous_action-seed1/commit/d27a3d5}{\footnotesize\texttt{d27a3d5}} \\
        \midrule

        \multirow{5}{*}{\footnotesize{Walker2d-v4}}
        & \includegraphics[height=0.8em]{media/emoji_pin.pdf}\blindhreftiny{https://huggingface.co/sdpkjc/Walker2d-v4-ppo_fix_continuous_action-seed4}{\footnotesize\texttt{Walker2d-v4-ppo\_fix\_continuous\_action-seed4}} & \footnotesize$4735.58 \pm 1183.56$ & \blindhreftiny{https://huggingface.co/sdpkjc/Walker2d-v4-ppo_fix_continuous_action-seed4/commit/9df90bd}{\footnotesize\texttt{9df90bd}} \\
        & \blindhreftiny{https://huggingface.co/sdpkjc/Walker2d-v4-ppo_fix_continuous_action-seed2}{\footnotesize\texttt{Walker2d-v4-ppo\_fix\_continuous\_action-seed2}} & \footnotesize$4057.75 \pm 1062.76$ & \blindhreftiny{https://huggingface.co/sdpkjc/Walker2d-v4-ppo_fix_continuous_action-seed2/commit/b25b341}{\footnotesize\texttt{b25b341}} \\
        & \blindhreftiny{https://huggingface.co/sdpkjc/Walker2d-v4-ppo_fix_continuous_action-seed3}{\footnotesize\texttt{Walker2d-v4-ppo\_fix\_continuous\_action-seed3}} & \footnotesize$3781.41 \pm 1202.34$ & \blindhreftiny{https://huggingface.co/sdpkjc/Walker2d-v4-ppo_fix_continuous_action-seed3/commit/907651a}{\footnotesize\texttt{907651a}} \\
        & \blindhreftiny{https://huggingface.co/sdpkjc/Walker2d-v4-ppo_fix_continuous_action-seed1}{\footnotesize\texttt{Walker2d-v4-ppo\_fix\_continuous\_action-seed1}} & \footnotesize$3357.25 \pm 1235.64$ & \blindhreftiny{https://huggingface.co/sdpkjc/Walker2d-v4-ppo_fix_continuous_action-seed1/commit/28a01f1}{\footnotesize\texttt{28a01f1}} \\
        & \blindhreftiny{https://huggingface.co/sdpkjc/Walker2d-v4-ppo_fix_continuous_action-seed5}{\footnotesize\texttt{Walker2d-v4-ppo\_fix\_continuous\_action-seed5}} & \footnotesize$2401.69 \pm 876.52$ & \blindhreftiny{https://huggingface.co/sdpkjc/Walker2d-v4-ppo_fix_continuous_action-seed5/commit/67e3c10}{\footnotesize\texttt{67e3c10}} \\
        \midrule
        
        \multirow{5}{*}{\footnotesize{HalfCheetah-v4}}
        & \includegraphics[height=0.8em]{media/emoji_pin.pdf}\blindhreftiny{https://huggingface.co/sdpkjc/HalfCheetah-v4-ppo_fix_continuous_action-seed1}{\footnotesize\texttt{HalfCheetah-v4-ppo\_fix\_continuous\_action-seed1}} & \footnotesize$4043.23 \pm 526.25$ & \blindhreftiny{https://huggingface.co/sdpkjc/HalfCheetah-v4-ppo_fix_continuous_action-seed1/commit/bc83fb6}{\footnotesize\texttt{bc83fb6}} \\
        & \blindhreftiny{https://huggingface.co/sdpkjc/HalfCheetah-v4-ppo_fix_continuous_action-seed4}{\footnotesize\texttt{HalfCheetah-v4-ppo\_fix\_continuous\_action-seed4}} & \footnotesize$2522.56 \pm 537.35$ & \blindhreftiny{https://huggingface.co/sdpkjc/HalfCheetah-v4-ppo_fix_continuous_action-seed4/commit/515348e}{\footnotesize\texttt{515348e}} \\
        & \blindhreftiny{https://huggingface.co/sdpkjc/HalfCheetah-v4-ppo_fix_continuous_action-seed2}{\footnotesize\texttt{HalfCheetah-v4-ppo\_fix\_continuous\_action-seed2}} & \footnotesize$1866.44 \pm 23.70$ & \blindhreftiny{https://huggingface.co/sdpkjc/HalfCheetah-v4-ppo_fix_continuous_action-seed2/commit/871ea55}{\footnotesize\texttt{871ea55}} \\
        & \blindhreftiny{https://huggingface.co/sdpkjc/HalfCheetah-v4-ppo_fix_continuous_action-seed5}{\footnotesize\texttt{HalfCheetah-v4-ppo\_fix\_continuous\_action-seed5}} & \footnotesize$1821.81 \pm 27.10$ & \blindhreftiny{https://huggingface.co/sdpkjc/HalfCheetah-v4-ppo_fix_continuous_action-seed5/commit/b007d7f}{\footnotesize\texttt{b007d7f}} \\
        & \blindhreftiny{https://huggingface.co/sdpkjc/HalfCheetah-v4-ppo_fix_continuous_action-seed3}{\footnotesize\texttt{HalfCheetah-v4-ppo\_fix\_continuous\_action-seed3}} & \footnotesize$1741.62 \pm 30.79$ & \blindhreftiny{https://huggingface.co/sdpkjc/HalfCheetah-v4-ppo_fix_continuous_action-seed3/commit/f696a66}{\footnotesize\texttt{f696a66}} \\
        \midrule
        
        \multirow{5}{*}{\footnotesize{Ant-v4}}
        & \includegraphics[height=0.8em]{media/emoji_pin.pdf}\blindhreftiny{https://huggingface.co/sdpkjc/Ant-v4-ppo_fix_continuous_action-seed2}{\footnotesize\texttt{Ant-v4-ppo\_fix\_continuous\_action-seed2}} & \footnotesize$3611.87 \pm 747.12$ & \blindhreftiny{https://huggingface.co/sdpkjc/Ant-v4-ppo_fix_continuous_action-seed2/commit/b88f77d}{\footnotesize\texttt{b88f77d}} \\
        & \blindhreftiny{https://huggingface.co/sdpkjc/Ant-v4-ppo_fix_continuous_action-seed3}{\footnotesize\texttt{Ant-v4-ppo\_fix\_continuous\_action-seed3}} & \footnotesize$2739.20 \pm 562.54$ & \blindhreftiny{https://huggingface.co/sdpkjc/Ant-v4-ppo_fix_continuous_action-seed3/commit/419360f}{\footnotesize\texttt{419360f}} \\
        & \blindhreftiny{https://huggingface.co/sdpkjc/Ant-v4-ppo_fix_continuous_action-seed4}{\footnotesize\texttt{Ant-v4-ppo\_fix\_continuous\_action-seed4}} & \footnotesize$2942.98 \pm 823.33$ & \blindhreftiny{https://huggingface.co/sdpkjc/Ant-v4-ppo_fix_continuous_action-seed4/commit/07048f2}{\footnotesize\texttt{07048f2}} \\
        & \blindhreftiny{https://huggingface.co/sdpkjc/Ant-v4-ppo_fix_continuous_action-seed5}{\footnotesize\texttt{Ant-v4-ppo\_fix\_continuous\_action-seed5}} & \footnotesize$2383.17 \pm 1044.23$ & \blindhreftiny{https://huggingface.co/sdpkjc/Ant-v4-ppo_fix_continuous_action-seed5/commit/3eec78a}{\footnotesize\texttt{3eec78a}} \\
        & \blindhreftiny{https://huggingface.co/sdpkjc/Ant-v4-ppo_fix_continuous_action-seed1}{\footnotesize\texttt{Ant-v4-ppo\_fix\_continuous\_action-seed1}} & \footnotesize$1866.34 \pm 766.40$ & \blindhreftiny{https://huggingface.co/sdpkjc/Ant-v4-ppo_fix_continuous_action-seed1/commit/be0d911}{\footnotesize\texttt{be0d911}} \\
        \midrule

        \multirow{5}{*}{\footnotesize{Swimmer-v4}}
        & \includegraphics[height=0.8em]{media/emoji_pin.pdf}\blindhreftiny{https://huggingface.co/sdpkjc/Swimmer-v4-ppo_fix_continuous_action-seed1}{\footnotesize\texttt{Swimmer-v4-ppo\_fix\_continuous\_action-seed1}} & \footnotesize$131.51 \pm 2.04$ & \blindhreftiny{https://huggingface.co/sdpkjc/Swimmer-v4-ppo_fix_continuous_action-seed1/commit/989c6ba}{\footnotesize\texttt{989c6ba}} \\
        & \blindhreftiny{https://huggingface.co/sdpkjc/Swimmer-v4-ppo_fix_continuous_action-seed4}{\footnotesize\texttt{Swimmer-v4-ppo\_fix\_continuous\_action-seed4}} & \footnotesize$119.79 \pm 2.48$ & \blindhreftiny{https://huggingface.co/sdpkjc/Swimmer-v4-ppo_fix_continuous_action-seed4/commit/5057fec}{\footnotesize\texttt{5057fec}} \\
        & \blindhreftiny{https://huggingface.co/sdpkjc/Swimmer-v4-ppo_fix_continuous_action-seed3}{\footnotesize\texttt{Swimmer-v4-ppo\_fix\_continuous\_action-seed3}} & \footnotesize$75.22 \pm 4.29$ & \blindhreftiny{https://huggingface.co/sdpkjc/Swimmer-v4-ppo_fix_continuous_action-seed3/commit/cc81c0e}{\footnotesize\texttt{cc81c0e}} \\
        & \blindhreftiny{https://huggingface.co/sdpkjc/Swimmer-v4-ppo_fix_continuous_action-seed2}{\footnotesize\texttt{Swimmer-v4-ppo\_fix\_continuous\_action-seed2}} & \footnotesize$63.36 \pm 1.08$ & \blindhreftiny{https://huggingface.co/sdpkjc/Swimmer-v4-ppo_fix_continuous_action-seed2/commit/63be675}{\footnotesize\texttt{63be675}} \\
        & \blindhreftiny{https://huggingface.co/sdpkjc/Swimmer-v4-ppo_fix_continuous_action-seed5}{\footnotesize\texttt{Swimmer-v4-ppo\_fix\_continuous\_action-seed5}} & \footnotesize$60.77 \pm 3.35$ & \blindhreftiny{https://huggingface.co/sdpkjc/Swimmer-v4-ppo_fix_continuous_action-seed5/commit/4435bb6}{\footnotesize\texttt{4435bb6}} \\
        \midrule

        \multirow{5}{*}{\footnotesize{Humanoid-v4}}
        & \includegraphics[height=0.8em]{media/emoji_pin.pdf}\blindhreftiny{https://huggingface.co/sdpkjc/Humanoid-v4-ppo_fix_continuous_action-seed4}{\footnotesize\texttt{Humanoid-v4-ppo\_fix\_continuous\_action-seed4}} & \footnotesize$704.90 \pm 153.81$ & \blindhreftiny{https://huggingface.co/sdpkjc/Humanoid-v4-ppo_fix_continuous_action-seed4/commit/83d57b0}{\footnotesize\texttt{83d57b0}} \\
        & \blindhreftiny{https://huggingface.co/sdpkjc/Humanoid-v4-ppo_fix_continuous_action-seed3}{\footnotesize\texttt{Humanoid-v4-ppo\_fix\_continuous\_action-seed3}} & \footnotesize$687.42 \pm 159.92$ & \blindhreftiny{https://huggingface.co/sdpkjc/Humanoid-v4-ppo_fix_continuous_action-seed3/commit/318aafa}{\footnotesize\texttt{318aafa}} \\
        & \blindhreftiny{https://huggingface.co/sdpkjc/Humanoid-v4-ppo_fix_continuous_action-seed2}{\footnotesize\texttt{Humanoid-v4-ppo\_fix\_continuous\_action-seed2}} & \footnotesize$645.69 \pm 143.65$ & \blindhreftiny{https://huggingface.co/sdpkjc/Humanoid-v4-ppo_fix_continuous_action-seed2/commit/b5dcc47}{\footnotesize\texttt{b5dcc47}} \\
        & \blindhreftiny{https://huggingface.co/sdpkjc/Humanoid-v4-ppo_fix_continuous_action-seed5}{\footnotesize\texttt{Humanoid-v4-ppo\_fix\_continuous\_action-seed5}} & \footnotesize$591.69 \pm 107.84$ & \blindhreftiny{https://huggingface.co/sdpkjc/Humanoid-v4-ppo_fix_continuous_action-seed5/commit/d08d91f}{\footnotesize\texttt{d08d91f}} \\
        & \blindhreftiny{https://huggingface.co/sdpkjc/Humanoid-v4-ppo_fix_continuous_action-seed1}{\footnotesize\texttt{Humanoid-v4-ppo\_fix\_continuous\_action-seed1}} & \footnotesize$640.32 \pm 171.90$ & \blindhreftiny{https://huggingface.co/sdpkjc/Humanoid-v4-ppo_fix_continuous_action-seed1/commit/e1edbff}{\footnotesize\texttt{e1edbff}} \\
        \bottomrule
    \end{tabular}
    \caption{PPO Models Evaluation Scores and Links}\label{tab:ppo-models}
\end{table}

\end{document}